\pdfoutput=1
\documentclass{achemso}
\usepackage[table,dvipsnames]{xcolor}
\usepackage{tikz}
\usepackage{float}
\usepackage{amsmath}
\usepackage{amssymb}
\usepackage{siunitx}
\usepackage{booktabs}
\usepackage{rotating}
\usepackage{multirow}
\usepackage{realboxes}
\usepackage{prettyref}
\usepackage{algorithm}
\usepackage{algpseudocode}
\usepackage{threeparttable}
\usepackage[utf8]{inputenc}
\usepackage[colorlinks,linkcolor=black,citecolor=blue]{hyperref}
\newrefformat{fig}{Figure~\ref{#1}}
\newrefformat{sec}{Section~\ref{#1}}
\newrefformat{table}{Table~\ref{#1}}
\newrefformat{eq}{Equation~(\ref{#1})}
\newrefformat{alg}{Algorithm~\ref{#1}}
\newcommand{\tabincell}[2]{\begin{tabular}{@{}#1@{}}#2\end{tabular}}
\definecolor{cff7f7f}{RGB}{255,127,127}
\definecolor{ccccc00}{RGB}{204,204,0}

\author{Nianze TAO}
\email{tao-nianze@hiroshima-u.ac.jp}
\author{Minori ABE}
\email{minoria@hiroshima-u.ac.jp}
\affiliation[Hiroshima University]
{Department of Chemistry, Graduate School of Advanced Science and Engineering, Hiroshima University, 1-3-1 Kagamiyama, Higashi-Hiroshima, Japan 739-8524}

\title{A Bayesian Flow Network Framework for Chemistry Tasks}

\abbreviations{BFN,MLP,AR,LM,DM,NLP,LLM,SMILES,SELFIES,SAFE,FCD,SOTA,MAE,RMSE,ROC-AUC,HTE,ADME}

\begin{tocentry}
    \includegraphics[width=8.47cm,height=4.67cm]{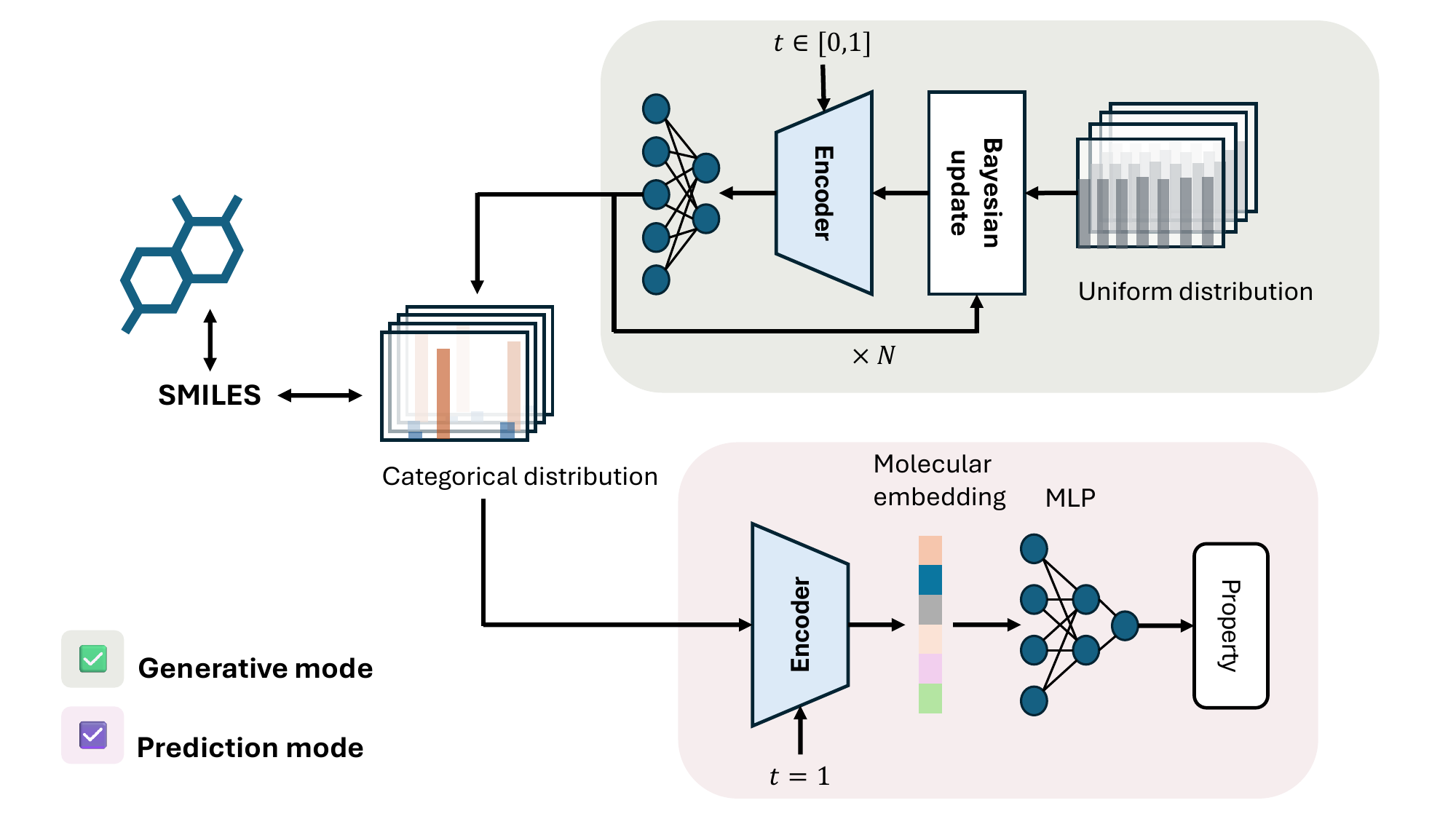}
\end{tocentry}

\begin{document}
	
	\begin{abstract}
		In this work, we introduce ChemBFN, a language model that handles chemistry tasks based on Bayesian flow networks working on discrete data. A new accuracy schedule is proposed to improve the sampling quality by significantly reducing the reconstruction loss. We show evidence that our method is appropriate for generating molecules with satisfied diversity even when a smaller number of sampling steps is used. A classifier-free guidance method is adapted for conditional generation. It is also worthwhile to point out that after generative training, our model can be fine-tuned on regression and classification tasks with the state-of-the-art performance, which opens the gate of building all-in-one models in a single module style. Our model has been open sourced at \url{https://github.com/Augus1999/bayesian-flow-network-for-chemistry}.
	\end{abstract}
	
	\section{Introduction}
	\par {\LARGE\bf A}utoregressive models (ARs) including SMILES-based or fragment-based models\cite{generating,rnn-guidelines,latentgan,molgpt,jtn-vae,guacamol,reinvent4,link-invent,release} that leverage the power of language models (LMs) and reinforcement learning\cite{reinvent4,link-invent,release} and graph-based models\cite{graphinvent,mcts,vgae-mcts,chemts,molrnn,dgm} coupled with advanced techniques such as Monte Carlo tree search\cite{mcts,vgae-mcts,chemts} have been proved their success in several \textit{de novo} design benchmarks\cite{moses,guacamol} consisted of drug-like molecules. The constraint of ARs, i.e., the number of sampling steps is the size of generated object, however, limits the potential of generating large molecules. Conversely, the recently emerging denoising-diffusion models\cite{ddpm} (DMs) offer a way to generate objects of any size within a fixed sequence of sampling process. However, it has been pointed out in the research of C. Vignac \textit{et al}\cite{digress} that SMILES-based models generally worked better than graph DMs even when a dedicatedly designed discrete diffusion method was applied.
	\par Bayesian flow networks\cite{bfn} (BFNs) are in a different category of generative models that decouple the sampling process with the size of generated objects as well. Different from DMs, BFNs directly work on the parameters of data distributions which naturally enable them to handle both continuous (including discretised) and discrete data without any data preprocessing or change of (mathematical) framework. Although the authors of BFN showed evidence in the original paper\cite{bfn} that BFN advantaged over discrete DMs on discrete data generating, e.g., text generation, the recent researches considering \textit{de novo} molecule design only successfully employed it on continuous and discretised data, e.g., 3D molecular conformation generation\cite{geobfn} rather than language-like representations such as SMILES\cite{smiles} or SELFIES\cite{selfies}. One potential reason discouraging the application to text generation is the lack of \textit{exact} analytical expression for the accuracy schedule $\beta(t)$, one critical component of BFNs, in the discrete case, while the speculated quadratic $\beta(t)$ in the original paper is, as admitted by the authors\cite{bfn}, suboptimal.
	\par In this paper, we introduce ChemBFN, a \underline{B}ayesian \underline{F}low \underline{N}etwork framework for \underline{Chem}istry tasks, that leverages our newly proposed accuracy schedule and transformer\cite{attention} encoder model to generate 1D language-like molecular representations e.g., SMILES and SELFIES. The experiments demonstrated that models with our accuracy schedule outperform those with the quadratic accuracy schedule. Besides, the generative training of BFN method can be a powerful pretraining strategy for downstream tasks in molecular property predictions, including regressions and classifications, and reaction yield predictions.
	
	\section{Methods}
    \subsection{Discrete Bayesian Flow Networks}
    \par A functional BFN is consisted of a neural network (NN) model that converts the \textit{input distribution} $\boldsymbol{p}_{I}(\boldsymbol{x}|\boldsymbol{\theta})$ into the \textit{output distribution} $\boldsymbol{p}_{O}(\boldsymbol{x}|\boldsymbol{\theta};t)$ and a Bayesian update process that updates the pervious input distribution to the current state according to a \textit{sender distribution} $\boldsymbol{p}_{S}(\boldsymbol{y}|\boldsymbol{x};\alpha)$, where $\boldsymbol{\theta}$ is the parameter of \textit{data} $\boldsymbol{x}$, and $\boldsymbol{y}$ is a sample of $\boldsymbol{x}$\cite{bfn}. The none-negative monotonic increasing function $\alpha$, namely accuracy rate, guides the sender distribution to moving to a more informative direction along with the time\cite{bfn}. Since $\alpha$ can be either continuous or discretised, a continuous accuracy schedule $\beta(t)$ is defined instead, which generates $\alpha$ as
    \begin{equation}
        \alpha = \left\{\begin{array}{ll}
           {\dfrac{d}{dt}\beta(t)}, & {\textrm{when $\alpha$ is continuous}}\\
           {\beta(t_{i}) - \beta(t_{i-1})}, & {\textrm{when $\alpha$ is discretised}}.
        \end{array}\right.
    \end{equation}
    In the discrete case, we have (1) all the distributions are K-class categorical distributions; (2) the sample is defined as $\boldsymbol{y}=\mathcal{N}(\alpha(K\boldsymbol{e_{x}}-\boldsymbol{1}),\alpha K\boldsymbol{I})$ when a Gaussian sampling is utilised, where $\boldsymbol{e_{x}}$ is the one-hot representation of data $\boldsymbol{x}$; (3) the Bayesian update function is defined as $h(\theta^{(d)},y^{(d)},\alpha)=e^{y^{(d)}}\theta^{(d)}/\sum_{k=1}^{K}e^{y^{(d)}_{k}}\theta^{(d)}_{k}$, where $\cdot^{(d)}$ is the $d^{th}$ parameter\cite{bfn}.
    \par During the training stage, a \textit{receiver distribution} $\boldsymbol{p}_{R}(\hat{\boldsymbol{y}}|\boldsymbol{\theta};t,\alpha)$ is drawn by sampling the output of NN with the same sampling method as sender distribution\cite{bfn}. The model is optimised by minimising the Kullback-Leibler divergence between the receiver distribution and the sender distribution, which is decoupled as a $n$-step loss ($L^{n}$) and a reconstruction loss ($L^{r}$) and only the first loss in practice is used\cite{bfn}. The limit case, i.e, continuous time loss, $L^{\infty}=\lim_{n\rightarrow\infty}L^{n}$ has been proved more efficient\cite{bfn}. During the sampling (generating) stage, since the receiver distribution has been trained to match the sender distribution, i.e., $\boldsymbol{p}_{R}(\hat{\boldsymbol{y}}|\boldsymbol{\theta};t,\alpha)\sim\boldsymbol{p}_{S}(\boldsymbol{y}|\boldsymbol{x};\alpha)$, the receiver distribution is used in Bayesian update process directly to update the input distribution (initialised as a uniform distribution over K categories) where a discretised $\alpha$ is employed.
    
	\subsection{Model Architecture}
	\par
\begin{figure}[H]
		  \centering
		  \includegraphics[width=0.8\textwidth]{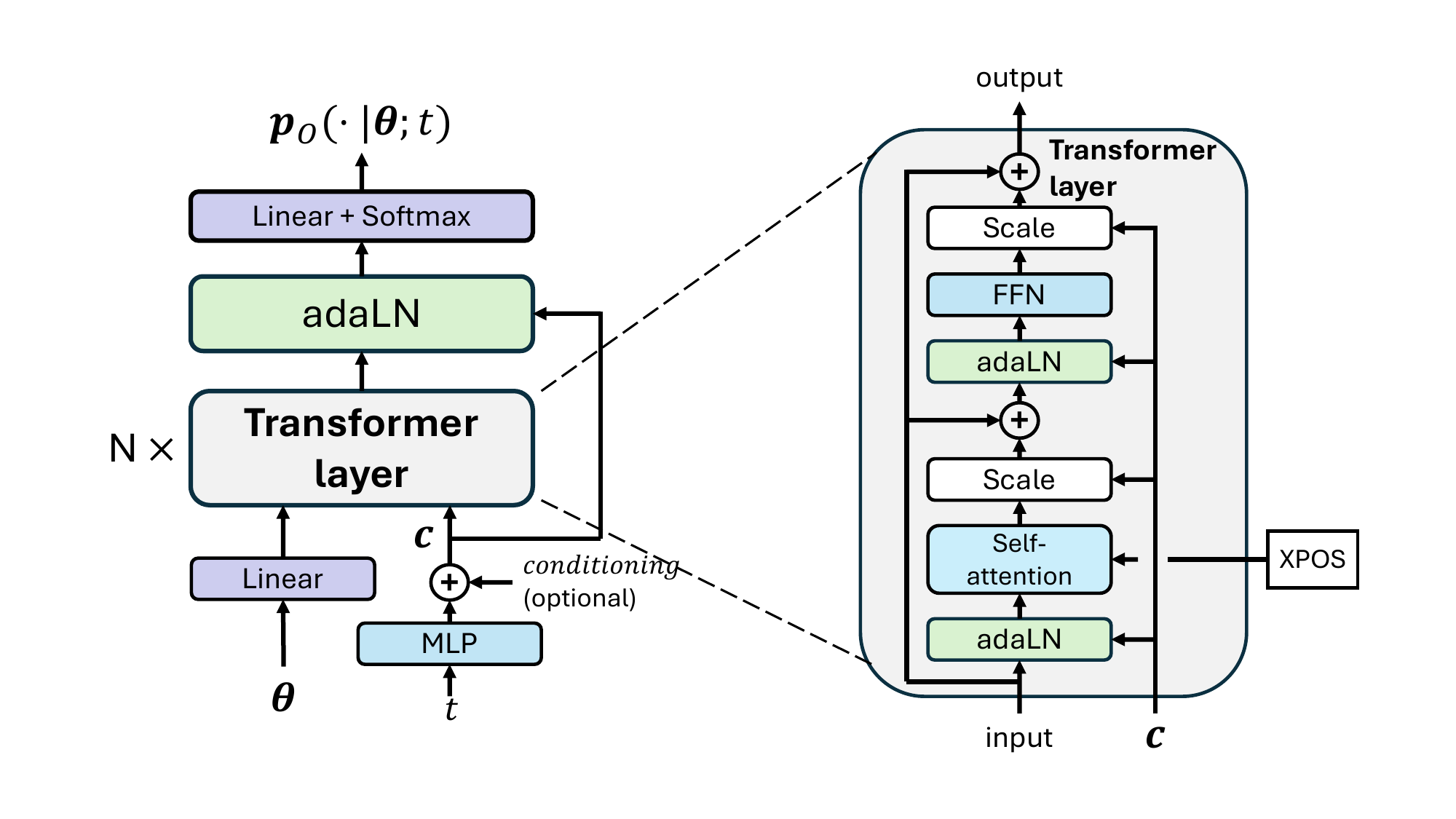}
		  \caption{Visualised scheme of our model. The architecture is inspired by DiT\protect\cite{dit}. The multi-head self-attention layers did not use causal masking which is the same as BERT\protect\cite{bert} while we replaced the commonly used positional embedding method (absolute positional embedding used in DiT, BERT and RoBERTa\protect\cite{roberta} models) with the novel {\small X{\normalsize P}OS}\protect\cite{xpos} variation of rotary positional embedding\protect\cite{roformer}. Note that each FFN (feed-forward network) layer adsorbs a dropout layer.}
		  \label{fig:model_architecture}
	\end{figure}

 Our model is an adaptation of DiT\cite{dit} model. The differences in our implementation include (1) the use of categorical distributions rather than image embeddings for input tokens because we are not dealing with images; (2) logits output that are then transformed into probabilities by softmax function; (3) the replacement of  activation function with SELU\cite{selu} function; (4) the use of a 2-layer multilayer perceptron (MLP) to form the time embedding since ``time'' in BFN is continuous from 0 to 1; (5) the employment of {\small X}P{\small OS}\cite{xpos} variation of rotary positional embedding\cite{roformer}. The architecture is shown in \prettyref{fig:model_architecture}.
	\par Following the notations of the BFN paper\cite{bfn}, the parameter of categorical distributions inputted into the neural network is denoted by $\boldsymbol{\theta} = (\theta^{(1)}, \theta^{(2)}, ..., \theta^{(D)}) \in [0,1]^{KD}$ ($K$ is the number of categories, $D$ is the number of input data, and $\theta^{(d)}$ is the $d^{th}$ parameter) and the output distribution at time step $t$ is denoted by $\boldsymbol{p}_{O}(\cdot|\boldsymbol{\theta};t) \in [0,1]^{KD}$. We denote the sum of time embedding vector and conditioning vector as $\boldsymbol{c}$. A null conditioning $\phi$ is equivalent to a zero vector $\boldsymbol{0}$.
	\par In each experiment described in the later text, we employed the same hyperparameters of the model except category number $K$ that depends on molecular representations. The 2-layer MLP with SELU activation has the shape of [1, 256, 512]. We employed 12 Transformer layers, of which had 8 attention heads each, with the attention temperature $\tau = \sqrt{2d_{h}}$ ($d_{h}$ is the feature number of each attention head)\cite{attention-temperature}. The dropout rate was 0.01 and the hidden feature number was 512. These settings lead to a total learnable parameters of the model of the magnitude of 54M.
	
	\subsection{A New Accuracy Schedule}
	\par In the case of BFN, an accuracy schedule function $\beta(t)$ drives the expectation of entropy of the input distribution $\mathbb{E}_{\boldsymbol{p}_{F}(\boldsymbol{\theta}|\boldsymbol{x};t)}H[\boldsymbol{p}_{I}(\boldsymbol{x}|\boldsymbol{\theta})]$ to decrease linearly with $t$, where $\boldsymbol{x}$ stands for the clear \textit{data},  $\boldsymbol{p}_{F}(\boldsymbol{\theta}|\boldsymbol{x};t)$ represents \textit{Bayesian flow distribution}, and $\boldsymbol{p}_{I}(\boldsymbol{x}|\boldsymbol{\theta})$ is the \textit{input distribution} as denoted in the original paper\cite{bfn}. The mathematical difficulty of deriving the expectation analytically in the discrete case compels us to speculate from intuition. The authors of BFN claimed that ``$\beta(t)=t^{2}\beta(1)$ was a reasonable approximation'', but disclosed later that finding a suitable value for the hyperparameter $\beta(1)$ was not an easy job\cite{bfn}.
	\par
\begin{figure}[H]
		  \centering
		  \includegraphics[width=0.8\textwidth]{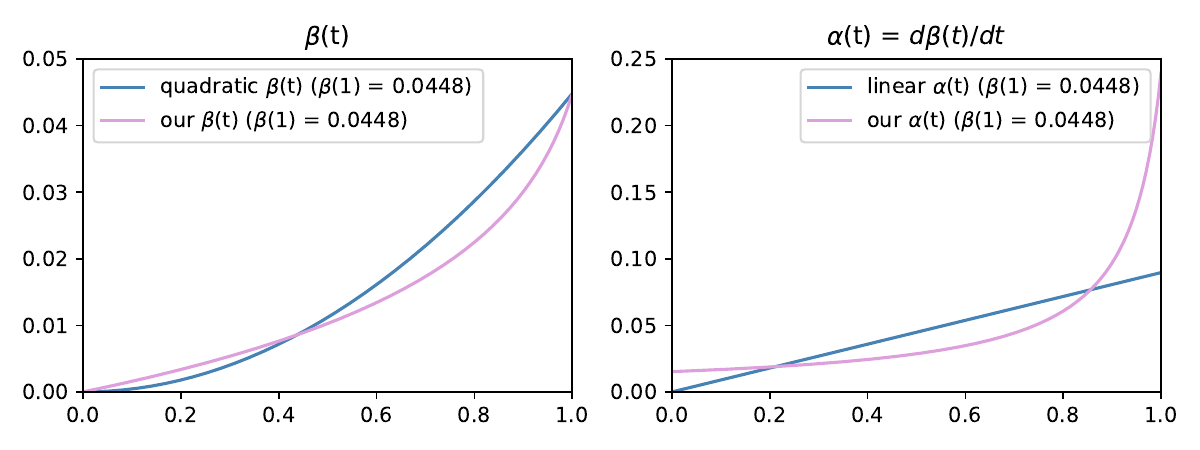}
		  \caption{Comparing our accuracy schedule with quadratic accuracy schedule initialised with the same value of $\beta(1)$. (Left) Accuracy schedules $\beta(t)$. (Right) The accuracy rates $\alpha(t)$. Note that our $\beta(t)$ does not deviate too much from quadratic one, yet the rate (derivative) differs substantially as $t$ goes to 1.}
		  \label{fig:visual_accuracy}
	\end{figure}

 Here, we give our estimation of $\beta(t)$. If we estimate the expected entropy of the input distribution (denoted as $E$ for short) as $E \sim f(K)e^{-\frac{K}{4}\beta(t)}$, then the relationship $E(t) = (1-t)E(0) + tE(1)$ that eliminates the unknown factor $f(K)$ gives us
	\begin{equation}
		\beta(t) = -\frac{4}{K}\ln{\left(1-t+te^{-\frac{K}{4}\beta(1)}\right)}
		\label{eq:beta_t}
	\end{equation}
	and the corresponding
	\begin{equation}
		\alpha(t) = \frac{d\beta}{dt} = \frac{4}{K}\frac{1-e^{-\frac{K}{4}\beta(1)}}{1-t+te^{-\frac{K}{4}\beta(1)}},
		\label{eq:alpha_t}
	\end{equation}
	where $\beta(1)$ is still a hyperparameter. \prettyref{eq:alpha_t} changes the continuous time loss $L^{\infty}$ to
	\begin{equation}
		L^{\infty}(\boldsymbol{x}) = \frac{K}{2}\mathbb{E}_{t\sim U(0,1),\boldsymbol{p}_{F}(\boldsymbol{\theta}|\boldsymbol{x};t)}\left(\alpha(t)\|\boldsymbol{e_{x}}-\hat{\boldsymbol{e}(\boldsymbol{\theta};t)}\|^{2}\right),
	\end{equation}
	where $\boldsymbol{e_{x}}$ is the one-hot representation of data $\boldsymbol{x}$ while $\hat{\boldsymbol{e}(\boldsymbol{\theta};t)}$ is the predicted categorical distribution of data $\boldsymbol{x}$ at time $t$. Note that when $\beta(1)$ is large, $\alpha(1)$ goes to extremely large. Therefore, we limit $\alpha(1) \leq 32\beta(1)$, from which
	\begin{equation}
		\beta(1)_{max} \approx 20.4054/K
	\end{equation}
	is obtained. An example of how our accuracy schedule looks different from original one is plotted in \prettyref{fig:visual_accuracy}. We shall show in the later experiments that our $\beta(t)$ in \prettyref{eq:beta_t} works better than quadratic ones.
		
	\subsection{Datasets and Benchmarks}
	\par Two benchmarks -- MOSES\cite{moses} and GuacaMol\cite{guacamol} -- were used to evaluate the generative performance, e.g., the similarity between generated molecules and training molecules, of ChemBFN. We reported the distribution-learning metrics of these benchmarks in \nameref{sec:experiments_and_results}. A summary of these metrics is in \prettyref{table:moses_guacamol_metrics}.
        \begin{table}[H]
		\centering
		\sisetup{table-format=6.0}
		\setlength\cmidrulewidth{\heavyrulewidth}
		\caption{A brief summary of used metrics of MOSES and GuacaMol benchmarks}
		\label{table:moses_guacamol_metrics}
		\begin{tabular}{ll}
			\toprule
			{Metrics} & {Description}\\
			\midrule
			\midrule
			{Valid} & {\small Fraction of valid molecules}\\
			{Unique} & {\small Fraction of unique molecules}\\
			{IntDiv$_{1}$ and IntDiv$_{2}$} & {\small Internal diversities}\\
			{Novelty} & {\small Fraction of generated unseen molecules compared with training data}\\
			{FCD} & {\small Fr\'echet ChemNet Distance\cite{fcd}}\\
			{SNN} & {\small Tanimoto similarity to a nearest neighbour}\\
			{Frag} & {\small BRICS fragment\cite{brics} cosine similarity}\\
			{Scaf} & {\small Bemis–Murcko scaffold\cite{bemis-murcko-scaffold} cosine similarity}\\
			{Filter} & {\small Fraction of molecules that fit pre-defined constructions}\\
			{KL Divergence} & {\small Kullback-Leibler divergence}\\
			\bottomrule
		\end{tabular}
	\end{table}
	\par The QM9\cite{qm9} dataset was employed to study the capability of conditional generation of our method. We randomly selected 110k molecules, before which 3054 invalid data were removed, with the triple $(\epsilon_{HOMO}, \epsilon_{LUMO},\Delta\epsilon_{HOMO-LUMO})$ as the conditioning label to form the training set.
	\par In order to evaluate the downstream performance, 40M unique SMILES and 190M unique SMILES strings were randomly selected from easily accessed ZINC15\cite{zinc15} database that formed two pretraining sets. The model trained on the 40M set was finetuned on several regression (ESOL, FreeSolv, Lipo, etc.) and classification (BBBP, BACE, ClinTox, HIV, etc.) tasks, including the subsets of widely used MoleculeNet\cite{moleculenet} benchmark. A brief description of used MoleculeNet tasks is in \prettyref{table:moleculenet_and_adme}. Each dataset was split into training/validation/testing sets in the ratio of 80/10/10 following the scaffold splitting method proposed in DeepChem\cite{deepchem} project. We reported ROC-AUC (area under receiver operating characteristic curve) for classification tasks and RMSE (root-mean squared error) for regression tasks in \nameref{sec:experiments_and_results}. In addition to the tasks of MoleculeNet, two less biased datasets -- the public ADME dataset published by C. Fang \textit{et al}\cite{adme} consisted of 6 dedicatedly collected absorption, distribution, metabolism, and excretion (ADME) \textit{in vitro} endpoints together with a Kinase inhibitor dataset prepared by J.Wu \textit{et al}\cite{kinase} that contains bioactivities of total 141,086 compounds for 354 kinases -- were employed to further benchmark our method in activity prediction. A brief summary of sub-tasks of ADME dataset is in \prettyref{table:moleculenet_and_adme}. For ADME dataset We employed the same split provided by C. Fang \textit{et al}\cite{adme}; for Kinase inhibitor dataset, we prepared a random split and a scaffold split (both had training/validation/testing = 80/10/10). The testing MAE, RMSE, Pearson's correlation coefficient (R value), and averaged ROC-AUC were reported in \nameref{sec:experiments_and_results}.
    \begin{table}[H]
		\centering
		\sisetup{table-format=6.0}
		\setlength\cmidrulewidth{\heavyrulewidth}
		\caption{A brief summary of used MoleculeNet and public ADME tasks}
		\label{table:moleculenet_and_adme}
        \scriptsize
		\begin{tabular}{lccl}
			\toprule
			{Name} & {\textnumero~molecules} & {\textnumero~tasks} & {Label}\\
			\midrule
			\midrule
			{ESOL} & {1,128} & {1} & {Aqueous solubility $log_{10}(S/{\rm mol\cdot L^{-1}})$}\\
			{FreeSolv} & {642} & {1} & {Experimental hydration free energy / kcal/mol}\\
			{Lipo} & {4,200} & {1} & {Octanol/water distribution coefficient $log_{10}D_{7.4}$}\\
                {HLM} & {3,087} & {1} & {Logarithm of human liver microsomal stability / $\rm mL\cdot min^{-1}\cdot kg^{-1}$}\\
                {RLM} & {3,054} & {1} & {Logarithm of rat liver microsomal stability / $\rm mL\cdot min^{-1}\cdot kg^{-1}$}\\
                {hPPB} & {1,808} & {1} & {Logarithm of human plasma protein binding (percent unbound)}\\
                {rPPB} & {884} & {1} & {Logarithm of rat plasma protein binding (percent unbound)}\\
                {MDR1-MDCK ER} & {2,642} & {1} & {$log_{10}(\textrm{MDR1-MDCK efflux ratio})$}\\
                {Solubility} & {2,173} & {1} & {Aqueous solubility $log_{10}(S/{\rm \mu g\cdot mL^{-1}})$ at PH = 6.8}\\
			\midrule
			{BBBP} & {2,039} & {1} & {If a compound penetrates the blood-brain barrier}\\
			{BACE} & {1,513} & {1} & {If a compound inhibits BACE-1 protein}\\
			\multirow{2}*{\tabincell{c}{ClinTox\\}} & \multirow{2}*{\tabincell{c}{1,478}} & \multirow{2}*{\tabincell{c}{2}} & {Task 1: if a drug has been approved by FDA}\\
			{} & {} & {} & {Task 2: if the same drug had toxicity during clinical trail}\\
			{HIV} & {41,127} & {1} & {If a compound is an HIV inhibitor}\\
			\bottomrule
		\end{tabular}
	\end{table}
	\par The USPTO-50k\cite{uspto-50k} dataset, Buchwald-Hartwig and Suzuki-Miyaura reaction yield datasets from high-throughput experiments (HTE) cleaned by P. Schwaller \textit{et al}\cite{yield-bert} were employed to train the model to predict reaction yield. USPTO-50k that contains 50k reactions mined from patents were used to pre-train the model while HTE data were used for fine-tuning. We report coefficient of determination (R$^{2}$ score) on testing sets in \nameref{sec:experiments_and_results}.
	\par AqSolDB\cite{aqsoldb}, a more challenging solubility dataset containing more species than ESOL, was used to investigate the effect of the size of pretraining data. A training/validation/testing (80/10/10) split was generated using scaffold splitting method. Testing MAE (mean absolute error) and RMSE were reported in \nameref{sec:experiments_and_results}.
	\par For SMILES representation, we developed a universal tokeniser that generates a fixed number (specifically $K=246$) of unique vocabulary for any collection of molecules. The similar strategy was not applicable to SELFIES strings, which were translated from SMILES via official \textit{selfies}\cite{selfies} package, hereby the vocabulary should be computed separately for each dataset and the category number $K$ varies. Note that we include three special tokens \colorbox{gray!15}{$\langle$start$\rangle$}, \colorbox{gray!15}{$\langle$end$\rangle$}, and \colorbox{gray!15}{$\langle$pad$\rangle$} in the vocabulary.
	
	\subsection{Fine-tuning Strategy}
        \begin{figure}[H]
		  \centering
		  \includegraphics[width=0.8\textwidth]{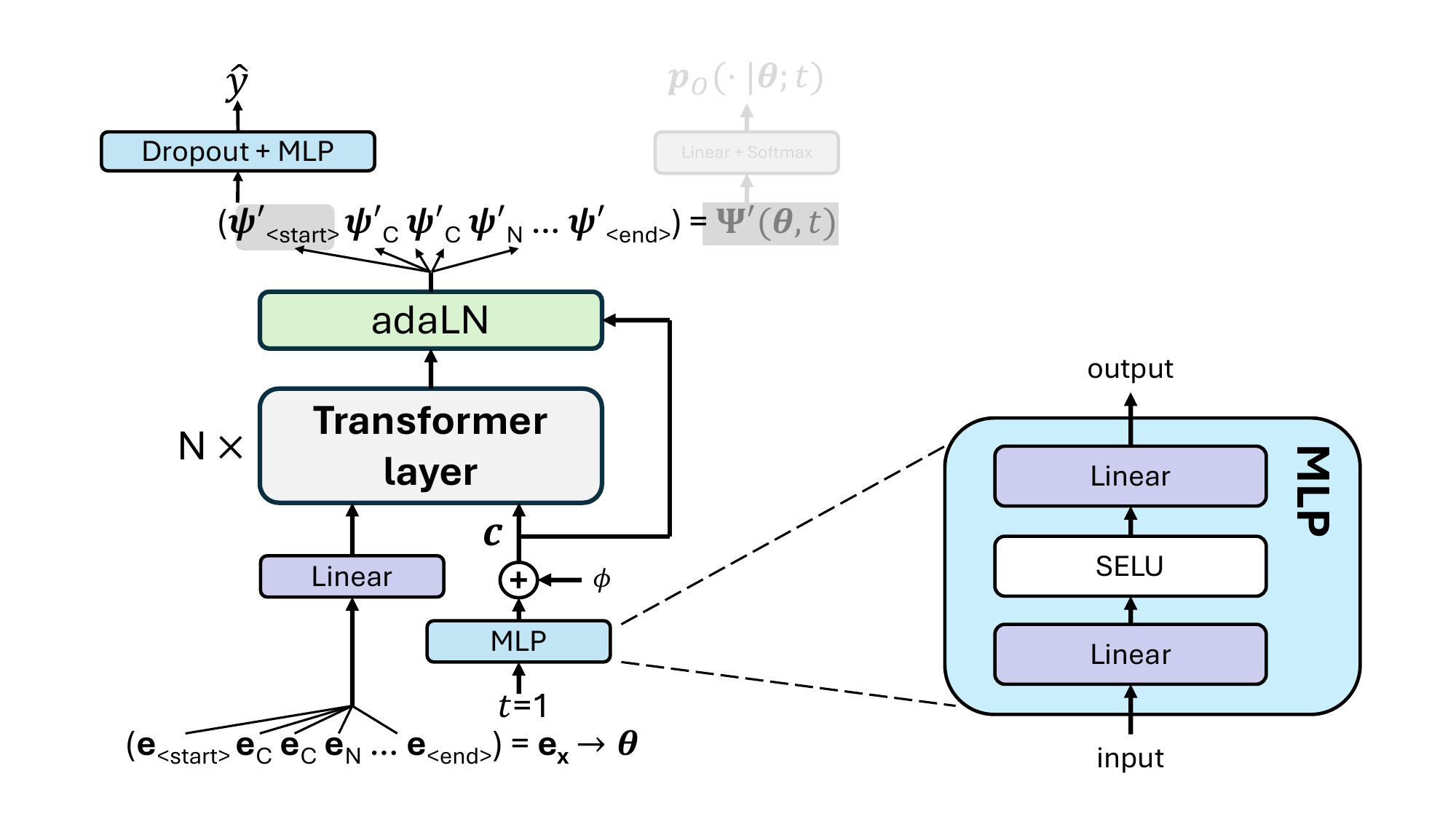}
		  \caption{The fine-tuning strategy of our model. The predicted label $\hat{y}\in\mathbb{R}^{n}$ is mapped by a MLP from the embedding of \colorbox{gray!15}{$\langle$start$\rangle$} token $\boldsymbol{\psi}'_{\langle{\rm start}\rangle}$ restricted by $t=1$. The MLP used here had 2 linear layers with a SELU activation function between them in a size of [512, 256, $n_{task}$]. Note that at prediction mode, the linear layer that maps latent vectors to output distributions is not activated; The conditioning is biased to null $\phi$; All \colorbox{gray!15}{$\langle$pad$\rangle$} tokens are masked out in attention.}
		  \label{fig:finetune_scheme}
	\end{figure}
	\par Similar to the strategy of ChemBERTa models\cite{chemberta,chemberta2}, the embedding, denoted as $\boldsymbol{\psi}'_{\langle{\rm start}\rangle}$, of \colorbox{gray!15}{$\langle$start$\rangle$} token at time $t=1$ was used as a fingerprint for downstream tasks. A 2-layer MLP absorbing a dropout layer is used as the prediction head. We replace the input distribution in generative mode with the one-hot representation of data (token), i.e., $\boldsymbol{\theta}\leftarrow \boldsymbol{e_{x}} = (\boldsymbol{e}_{\langle{\rm start}\rangle},...,\boldsymbol{e}_{\langle{\rm end}\rangle})\in\{0,1\}^{KD}$ in this stage. A visualised scheme is in \prettyref{fig:finetune_scheme}.
	
	\section{Experiments and Results}
        \label{sec:experiments_and_results}
	\subsection{Unconditional Generation}
	\begin{figure}[H]
		\centering
		\includegraphics[width=0.9\linewidth]{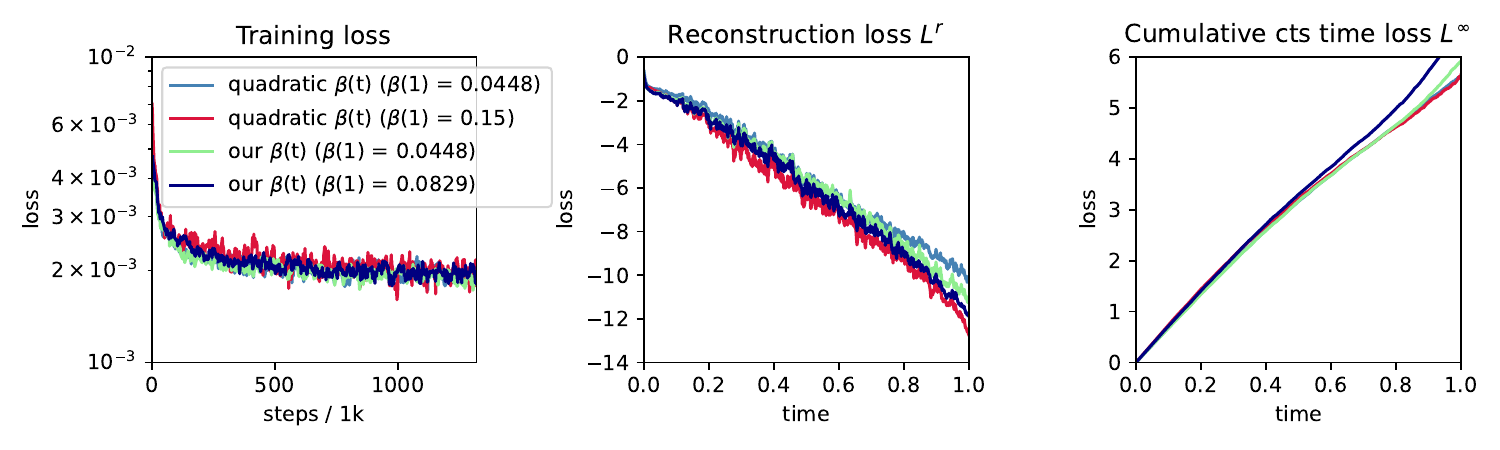}
		\caption{Visualisation of the impact on training loss, reconstruction loss $L^{r}$ and continuous (cts) time loss $L^{\infty}$ of different accuracy schedules with different values of $\beta(1)$. $L^{r}$ and $L^{\infty}$ were computed on 1k discretised steps after training.}
		\label{fig:visual_loss}
	\end{figure}
	\par We first evaluate the effect of different $\beta(t)$ with different values of $\beta(1)$ using MOSES dataset. We reported the validity, FCD on scaffold set, SNN on scaffold set, Frag on scaffold set, Scaf on scaffold set, Filters, and Novelty scores computed by MOSES program in \prettyref{table:accuracy_effect} together with reconstruction loss $L^{r} = -\mathbb{E}_{\boldsymbol{p}_{F}(\boldsymbol{\theta}|\mathbf{x};t)}\ln{\boldsymbol{p}_{O}(\mathbf{x}|\boldsymbol{\theta};t)}$ and continuous time loss $L^{\infty}$ in \prettyref{fig:visual_loss}. It is clear that raising $\beta(1)$ in both quadratic and our schedules did not have obvious influence on training loss but lowered $L^{r}$, while our schedule lead to a lower loss when $\beta(1)$ was the same. The effect on $L^{\infty}$ was subtle. However, after we calculated the $R^{2}$ values of the cumulative $L^{\infty}$ curves, we found that while using quadratic $\beta(t)$ the curve became more distorted when $\beta(1)$ was larger ($R^{2}|_{\beta(1)=0.0448}=0.995$ while $R^{2}|_{\beta(1)=0.15}=0.992$); After switching to our $\beta(t)$ the curves were more linear (i.e., $L^{\infty}$ was more uniform) and the linearity was not affected by the value of $\beta(1)$ ($R^{2}|_{\beta(1)=0.0448}=R^{2}|_{\beta(1)=0.0829}=0.997$). The metrics in \prettyref{table:accuracy_effect} provide more quantitative evidences that our $\beta(t)$ is more optimal. It is notable that a larger $\beta(1)$ value usually result to better scores. Therefore, we conclude here that our proposed $\beta(t)$ with $\beta(1)=\beta(1)_{max}=20.4054/K$ is a more optimal solution in discrete BFNs.
	\begin{table}[H]
        \centering
        \sisetup{table-format=6.0}
		\setlength\cmidrulewidth{\heavyrulewidth}
		\caption{Comparing scores of MOSES benchmark when varying $\beta(1)$ value of different accuracy schedules\textsuperscript{\emph{a}}}
        \label{table:accuracy_effect}
        \Makebox[\textwidth][c]{
        \begin{threeparttable}
        \scriptsize
		\begin{tabular}{cl|ccccccc}
			\toprule
			\multicolumn{2}{c}{$\beta(1)$} & {Valid $\uparrow$} & {FCD $\downarrow$} & {SNN $\uparrow$} & {Frag $\uparrow$} & {Scaf $\uparrow$} & {Filters $\uparrow$} & {Novelty $\uparrow$}\\
			\midrule
			\midrule
			\multirow{3}*{\tabincell{c}{\begin{turn}{90}\scriptsize quad\end{turn}\\}} & {0.15} & {0.893 {\tiny $\pm$ 0.001}} & {3.438 {\tiny $\pm$ 0.034}} & {0.559 {\tiny $\pm$ 0.000}} & {0.985 {\tiny $\pm$ 0.000}} & {0.095 {\tiny $\pm$ 0.001}} & {0.982 {\tiny $\pm$ 0.000}} & {{\bf 0.900} {\tiny $\pm$ 0.002}}\\
			{} & {0.0829} & {0.895 {\tiny $\pm$ 0.001}} & {3.772 {\tiny $\pm$ 0.012}} & {0.551 {\tiny $\pm$ 0.001}} & {0.984 {\tiny $\pm$ 0.001}} & {{\bf 0.096} {\tiny $\pm$ 0.006}} & {0.985 {\tiny $\pm$ 0.001}} & {{\bf 0.900} {\tiny $\pm$ 0.002}}\\
			{} & {0.0448} & {0.899 {\tiny $\pm$ 0.003}} & {3.902 {\tiny $\pm$ 0.045}} & {0.561 {\tiny $\pm$ 0.000}} & {0.988 {\tiny $\pm$ 0.000}} & {0.089 {\tiny $\pm$ 0.006}} & {0.986 {\tiny $\pm$ 0.001}} & {0.887 {\tiny $\pm$ 0.003}}\\
			\midrule
			\multirow{2}*{\tabincell{c}{\begin{turn}{90}\scriptsize ours\end{turn}\\}} & {0.0829} & {{\bf 0.900} {\tiny $\pm$ 0.001}} & {{\bf 2.731} {\tiny $\pm$ 0.015}} & {0.563 {\tiny $\pm$ 0.000}} & {{\bf 0.990} {\tiny $\pm$ 0.000}} & {0.091 {\tiny $\pm$ 0.004}} & {{\bf 0.987} {\tiny $\pm$ 0.001}} & {0.886 {\tiny $\pm$ 0.000}}\\
			{} & {0.0448} & {{\bf 0.900} {\tiny $\pm$ 0.001}} & {3.580 {\tiny $\pm$ 0.008}} & {{\bf 0.568} {\tiny $\pm$ 0.000}} & {0.987 {\tiny $\pm$ 0.000}} & {0.075 {\tiny $\pm$ 0.006}} & {{\bf 0.987} {\tiny $\pm$ 0.000}} & {0.877 {\tiny $\pm$ 0.001}}\\
			\bottomrule
		\end{tabular}
        \small
        \begin{tablenotes}
            \item [a] $\uparrow$ indicates that the higher is better and $\downarrow$ stands for the contrary. The best results are in \textbf{\textit{bold}}. We used a sampling step of 1k.
        \end{tablenotes} 
        \end{threeparttable}}
	\end{table}
	\par In the above experiments, we used a dynamic padding strategy, i.e., each batch were padded to the maximum length of that batch, to reduce the training time. In the following experiments, global padding strategy, i.e., padding all batches to a global maximum length, was employed to compare with dynamic strategy on both MOSES and GuacaMol benchmarks. The results were summarised in \prettyref{table:pad_strategy}. We found that the global padding method benefited the performance. In the following experiment, we therefore employed the global padding method in generative tasks.
	\begin{table}[H]
		\centering
		\sisetup{table-format=6.0}
		\setlength\cmidrulewidth{\heavyrulewidth}
		\caption{Scores of MOSES and GuacaMol benchmarks when different padding strategies were used during training\textsuperscript{\emph{a}}}
        \label{table:pad_strategy}
        \Makebox[\textwidth][c]{
        \begin{threeparttable}
        \scriptsize
		\begin{tabular}{c|ccccccc}
			\toprule
			\multirow{2}*{\tabincell{c}{\tiny Strategy\\}} & \multicolumn{7}{c}{MOSES}\\
			{} & {Valid $\uparrow$} & {FCD $\downarrow$} & {SNN $\uparrow$} & {Frag $\uparrow$} & {Scaf $\uparrow$} & {Filters $\uparrow$} & {Novelty $\uparrow$}\\
			\midrule
			{\tiny dynamic} & {0.900 {\tiny $\pm$ 0.001}} & {2.731 {\tiny $\pm$ 0.015}} & {0.563 {\tiny $\pm$ 0.000}} & {{\bf 0.990}  {\tiny $\pm$ 0.000}} & {0.091 {\tiny $\pm$ 0.004}} & {{\bf 0.987} {\tiny $\pm$ 0.001}} & {{\bf 0.886} {\tiny $\pm$ 0.000}}\\
			{\tiny global} & {{\bf 0.916} {\tiny $\pm$ 0.001}} & {{\bf 2.730} {\tiny $\pm$ 0.014}} & {{\bf 0.565} {\tiny $\pm$ 0.001}} & {{\bf 0.990} {\tiny $\pm$ 0.000}} & {{\bf 0.094} {\tiny $\pm$ 0.002}} & {{\bf 0.987} {\tiny $\pm$ 0.001}} & {0.880 {\tiny $\pm$ 0.002}}\\
			\midrule
			\midrule
			{} & \multicolumn{7}{c}{GuacaMol}\\
			{} & \multicolumn{2}{c}{Valid $\uparrow$} & {Unique $\uparrow$} & {Novelty $\uparrow$} & {KL Divergence $\uparrow$} & \multicolumn{2}{c}{FCD $\uparrow$}\\
			\midrule
			{\tiny dynamic} & \multicolumn{2}{c}{0.799 {\tiny $\pm$ 0.003}} & {0.815 {\tiny $\pm$ 0.002}} & {0.975 {\tiny $\pm$ 0.000}} & {{\bf 0.810} {\tiny $\pm$ 0.001}} & \multicolumn{2}{c}{0.370 {\tiny $\pm$ 0.003}}\\
			{\tiny global} & \multicolumn{2}{c}{{\bf 0.807} {\tiny $\pm$ 0.003}} & {{\bf 0.818} {\tiny $\pm$ 0.001}} & {{\bf 0.975} {\tiny $\pm$ 0.001}} & {0.808 {\tiny $\pm$ 0.010}} & \multicolumn{2}{c}{{\bf 0.399} {\tiny $\pm$ 0.002}}\\
			\bottomrule
		\end{tabular}
        \small
        \begin{tablenotes}
            \item [a] $\uparrow$ for higher is better and $\downarrow$ for contrary. The best results are in \textbf{\textit{bold}}. We used a sampling step of 1k.
        \end{tablenotes}
        \end{threeparttable}}
	\end{table}
	\par Finally, we trained models applying the above optimal settings (i.e., $\beta(1)=20.4054/K$ and global padding) on MOSES and GuacaMol datasets. Both SMILES and SELFIES versions were implemented. The comparison with published state-of-the-art (SOTA) models\cite{jtn-vae,latentgan,graphinvent,molgpt,vgae-mcts,digress,guacamol} are summarised in \prettyref{table:moses_result}, \prettyref{table:moses_sf_result}, and \prettyref{table:guacamol_result}. We found that (1) except FCD, metrics of both SMILES version and SELFIES version were close to SOTA performance. (2) number of sampling step as expected affected the validity of generated molecules (for SMILES version only because SELFIES \textit{always} gives valid molecules\cite{selfies}), but dropping from 1k steps to 100 steps did not degrade the performance a lot. If lower validity is acceptable, only sampling 10 steps significantly reduce the computational time without much impact on other qualities. Larger FCD (in the term of GuacaMol is lower FCD score where FCD score = $e^{-0.2 \rm FCD}$) is a hint that BFNs learn the grammar of molecules rather than the way of combining characters within the dataset.
	\begin{table}[H]
		\centering
		\sisetup{table-format=6.0}
		\setlength\cmidrulewidth{\heavyrulewidth}
		\caption{Testing metrics on MOSES test set compared with SOTA models\textsuperscript{\emph{a}}}
        \label{table:moses_result}
        \begin{threeparttable}
        \scriptsize
		\begin{tabular}{cl|cccccc}
			\toprule
			\multicolumn{2}{c}{Method} & {Valid $\uparrow$} & {Unique@1k $\uparrow$} & {Unique@10k $\uparrow$} & {IntDiv$_{1}$ $\uparrow$} & {IntDiv$_{2}$ $\uparrow$} & {Novelty $\uparrow$}\\
			\midrule
			\midrule
			\multirow{4}*{\tabincell{c}{\begin{turn}{90}\scriptsize ARs\end{turn}\\}} & {JTN-VAE\cite{jtn-vae}} & {{\bf 1.0} {\tiny $\pm$ 0.0}} & {{\bf 1.0} {\tiny $\pm$ 0.0}} & {{\bf 1.0} {\tiny $\pm$ 0.0}} & {0.855 {\tiny $\pm$ 0.003}} & {0.850 {\tiny $\pm$ 0.004}} & {0.913 {\tiny $\pm$ 0.006}}\\
			{} & {LatentGAN\cite{latentgan}} & {0.897 {\tiny $\pm$ 0.003}} & {{\bf 1.0} {\tiny $\pm$ 0.0}} & {0.997 {\tiny $\pm$ 0.000}} & {0.857 {\tiny $\pm$ 0.001}} & {0.851 {\tiny $\pm$ 0.001}} & {0.950 {\tiny $\pm$ 0.001}}\\
			{} & {GraphINVENT\cite{graphinvent}} & {0.964} & {\bf 1.0} & {0.998} & {0.857} & {0.851} & {--}\\
			{} & {MolGPT\cite{molgpt}} & {0.994} & {--} & {\bf 1.0} & {0.857} & {0.851} & {0.797}\\
			\midrule
			\midrule
			\multirow{2}*{\tabincell{c}{\begin{turn}{90}\scriptsize DMs\end{turn}\\}} & \multirow{2}*{\tabincell{c}{DiGress\cite{digress}\\}} & \multirow{2}*{\tabincell{c}{0.857\\}} & \multirow{2}*{\tabincell{c}{--\\}} & \multirow{2}*{\tabincell{c}{\bf 1.0\\}} & \multirow{2}*{\tabincell{c}{--\\}} & \multirow{2}*{\tabincell{c}{--\\}} & \multirow{2}*{\tabincell{c}{0.950\\}}\\
			{} & {} & {} & {} & {} & {} & {} & {}\\
			\midrule
			\midrule
			\multirow{6}*{\tabincell{c}{\begin{turn}{90}\scriptsize BFNs\end{turn}\\}} & {ChemBFN$_{10}$} & {0.835 {\tiny $\pm$ 0.003}} & {{\bf 1.0} {\tiny $\pm$ 0.0}} & {0.999 {\tiny $\pm$ 0.000}} & {0.851 {\tiny $\pm$ 0.000}} & {0.844 {\tiny $\pm$ 0.000}} & {0.921 {\tiny $\pm$ 0.002}}\\
			{} & {ChemBFN$_{100}$} & {0.911 {\tiny $\pm$ 0.002}} & {{\bf 1.0} {\tiny $\pm$ 0.0}} & {0.998 {\tiny $\pm$ 0.000}} & {0.837 {\tiny $\pm$ 0.000}} & {0.831 {\tiny $\pm$ 0.000}} & {0.884 {\tiny $\pm$ 0.002}}\\
			{} & {ChemBFN$_{\rm 1k}$} & {0.916 {\tiny $\pm$ 0.001}} & {{\bf 1.0} {\tiny $\pm$ 0.0}} & {0.998 {\tiny $\pm$ 0.000}} & {0.836 {\tiny $\pm$ 0.000}} & {0.830 {\tiny $\pm$ 0.000}} & {0.880 {\tiny $\pm$ 0.002}}\\
			{} & {ChemBFN$_{10}^{*}$} & {{\bf 1.0} {\tiny $\pm$ 0.0}} & {{\bf 1.0} {\tiny $\pm$ 0.0}} & {{\bf 1.0} {\tiny $\pm$ 0.0}} & {{\bf 0.860} {\tiny $\pm$ 0.000}} & {{\bf 0.855} {\tiny $\pm$ 0.000}} & {{\bf 0.991} {\tiny $\pm$ 0.000}}\\
			{} & {ChemBFN$_{100}^{*}$} & {{\bf 1.0} {\tiny $\pm$ 0.0}} & {{\bf 1.0} {\tiny $\pm$ 0.0}} & {{\bf 1.0} {\tiny $\pm$ 0.0}} & {0.848 {\tiny $\pm$ 0.000}} & {0.842 {\tiny $\pm$ 0.000}} & {0.947 {\tiny $\pm$ 0.001}}\\
			{} & {ChemBFN$_{\rm 1k}^{*}$} & {{\bf 1.0} {\tiny $\pm$ 0.0}} & {{\bf 1.0} {\tiny $\pm$ 0.0}} & {{\bf 1.0} {\tiny $\pm$ 0.0}} & {0.847 {\tiny $\pm$ 0.000}} & {0.841 {\tiny $\pm$ 0.000}} & {0.940 {\tiny $\pm$ 0.001}}\\
			\bottomrule
		\end{tabular}
        \small
        \begin{tablenotes}
            \item [a] The metrics of all other models were copied from the original paper. $\uparrow$ for the higher is better. (10, 100, 1k) are the number of sampling steps. * for SELFIES version. The best results are in \textbf{\textit{bold}}.
        \end{tablenotes}
        \end{threeparttable}
	\end{table}
	\begin{table}[H]
		\centering
		\sisetup{table-format=6.0}
		\setlength\cmidrulewidth{\heavyrulewidth}
		\caption{Metrics on MOSES {\bf scaffold} test set\textsuperscript{\emph{a}}}
        \label{table:moses_sf_result}
        \begin{threeparttable}
        \scriptsize
		\begin{tabular}{cl|ccccc}
			\toprule
			\multicolumn{2}{c}{Method} & {FCD $\downarrow$} & {SNN $\uparrow$} & {Frag $\uparrow$} & {Scaff $\uparrow$} & {Filters $\uparrow$}\\
			\midrule
			\midrule
			\multirow{3}*{\tabincell{c}{\begin{turn}{90}\scriptsize ARs\end{turn}\\}} & {JTN-VAE\cite{jtn-vae}} & {0.938 {\tiny $\pm$ 0.053}} & {0.519 {\tiny $\pm$ 0.007}} & {0.995 {\tiny $\pm$ 0.000}} & {0.101 {\tiny $\pm$ 0.011}} & {0.976 {\tiny $\pm$ 0.002}}\\
			{} & {LatentGAN\cite{latentgan}} & {{\bf 0.828} {\tiny $\pm$ 0.012}} & {0.513 {\tiny $\pm$ 0.000}} & {{\bf 0.997} {\tiny $\pm$ 0.001}} & {0.107 {\tiny $\pm$ 0.010}} & {0.974 {\tiny $\pm$ 0.001}}\\
			{} & {GraphINVENT\cite{graphinvent}} & {1.223} & {0.539} & {0.986} & {0.127} & {0.950}\\
			\midrule
			\midrule
			\multirow{2}*{\tabincell{c}{\begin{turn}{90}\scriptsize DMs\end{turn}\\}} & \multirow{2}*{\tabincell{c}{DiGress\cite{digress}\\}} & \multirow{2}*{\tabincell{c}{1.19\\}} & \multirow{2}*{\tabincell{c}{0.52\\}} & \multirow{2}*{\tabincell{c}{--\\}} & \multirow{2}*{\tabincell{c}{\bf 0.148\\}} & \multirow{2}*{\tabincell{c}{0.971\\}}\\
			{} & {} & {} & {} & {} & {} & {}\\
			\midrule
			\midrule
			\multirow{6}*{\tabincell{c}{\begin{turn}{90}\scriptsize BFNs\end{turn}\\}} & {ChemBFN$_{10}$} & {2.768 {\tiny $\pm$ {0.035}}} & {0.533 {\tiny $\pm$ {0.000}}} & {0.988 {\tiny $\pm$ {0.000}}} & {0.145 {\tiny $\pm$ {0.004}}} & {0.976 {\tiny $\pm$ {0.001}}}\\
			{} & {ChemBFN$_{100}$} & {2.604 {\tiny $\pm$ 0.040}} & {0.562 {\tiny $\pm$ 0.001}} & {0.991 {\tiny $\pm$ 0.000}} & {0.103 {\tiny $\pm$ 0.005}} & {0.985 {\tiny $\pm$ 0.001}}\\
			{} & {ChemBFN$_{\rm 1k}$} & {2.730 {\tiny $\pm$ 0.014}} & {{\bf 0.565} {\tiny $\pm$ 0.001}} & {0.990 {\tiny $\pm$ 0.000}} & {0.094 {\tiny $\pm$ 0.002}} & {{\bf 0.987} {\tiny $\pm$ 0.001}}\\
			{} & {ChemBFN$_{10}^{*}$} & {11.79 {\tiny $\pm$ 0.09}} & {0.422 {\tiny $\pm$ 0.001}} & {0.965 {\tiny $\pm$ 0.001}} & {0.118 {\tiny $\pm$ 0.016}} & {0.806 {\tiny $\pm$ 0.001}}\\
			{} & {ChemBFN$_{100}^{*}$} & {4.802 {\tiny $\pm$ 0.045}} & {0.517 {\tiny $\pm$ 0.000}} & {0.976 {\tiny $\pm$ 0.001}} & {0.141 {\tiny $\pm$ 0.008}} & {0.955 {\tiny $\pm$ 0.001}}\\
			{} & {ChemBFN$_{\rm 1k}^{*}$} & {4.473 {\tiny $\pm$ 0.058}} & {0.524 {\tiny $\pm$ 0.001}} & {0.976 {\tiny $\pm$ 0.000}} & {0.141 {\tiny $\pm$ 0.008}} & {0.962 {\tiny $\pm$ 0.001}}\\
			\bottomrule
		\end{tabular}
        \small
        \begin{tablenotes}
            \item [a] Settings are the same as \protect\prettyref{table:moses_result} while $\downarrow$ for the lower is better.
        \end{tablenotes}
        \end{threeparttable}
	\end{table}
	\begin{table}[H]
		\centering
		\sisetup{table-format=6.0}
		\setlength\cmidrulewidth{\heavyrulewidth}
		\caption{Testing metrics on GuacaMol distribution-learning tasks\textsuperscript{\emph{a}}}
        \label{table:guacamol_result}
        \begin{threeparttable}
        \scriptsize
		\begin{tabular}{cl|ccccc}
			\toprule
			\multicolumn{2}{c}{Method} & {Valid $\uparrow$} & {Unique $\uparrow$} & {Novelty $\uparrow$} & {KL Divergence $\uparrow$} & {FCD $\uparrow$}\\
			\midrule
			\midrule
			\multirow{3}*{\tabincell{c}{\begin{turn}{90}\scriptsize ARs\end{turn}\\}} & {MolGPT\cite{molgpt}} & {0.981} & {0.998} & {\bf 1.0} & {\bf 0.992} & {0.907}\\
			{} & {SMILES LSTM\cite{guacamol}} & {0.959} & {\bf 1.0} & {0.912} & {0.991} & {\bf 0.913}\\
			{} & {VGAE-MCTS\cite{vgae-mcts}} & {\bf 1.0} & {\bf 1.0} & {\bf 1.0} & {0.659} & {0.009}\\
			\midrule
			\midrule
			\multirow{2}*{\tabincell{c}{\begin{turn}{90}\scriptsize DMs\end{turn}\\}} & \multirow{2}*{\tabincell{c}{DiGress\cite{digress}\\}} & \multirow{2}*{\tabincell{c}{0.852\\}} & \multirow{2}*{\tabincell{c}{\bf 1.0\\}} & \multirow{2}*{\tabincell{c}{0.999\\}} & \multirow{2}*{\tabincell{c}{0.929\\}} & \multirow{2}*{\tabincell{c}{0.680\\}}\\
			{} & {} & {} & {} & {} & {} & {}\\
			\midrule
			\midrule
			\multirow{4}*{\tabincell{c}{\begin{turn}{90}\scriptsize BFNs\end{turn}\\}} & {ChemBFN$_{\rm 1k}$} & {0.807 {\tiny $\pm$ 0.003}} & {0.818 {\tiny $\pm$ 0.001}} & {0.975 {\tiny $\pm$ 0.001}} & {0.808 {\tiny $\pm$ 0.010}} & {0.399 {\tiny $\pm$ 0.002}}\\
			{} & {ChemBFN$_{10}^{*}$} & {{\bf 1.0} {\tiny $\pm$ 0.0}} & {0.853 {\tiny $\pm$ 0.002}} & {{\bf 1.0} {\tiny $\pm$ 0.0}} & {0.451 {\tiny $\pm$ 0.001}} & {0.000 {\tiny $\pm$ 0.000}}\\
			{} & {ChemBFN$_{100}^{*}$} & {{\bf 1.0} {\tiny $\pm$ 0.0}} & {0.846 {\tiny $\pm$ 0.003}} & {0.994 {\tiny $\pm$ 0.000}} & {0.803 {\tiny $\pm$ 0.003}} & {0.110 {\tiny $\pm$ 0.003}}\\
			{} & {ChemBFN$_{\rm 1k}^{*}$} & {{\bf 1.0} {\tiny $\pm$ 0.0}} & {0.850 {\tiny $\pm$ 0.003}} & {0.994 {\tiny $\pm$ 0.001}} & {0.811 {\tiny $\pm$ 0.002}} & {0.142 {\tiny $\pm$ 0.003}}\\
			\bottomrule
		\end{tabular}
        \small
        \begin{tablenotes}
            \item [a] Settings are the same as \protect\prettyref{table:moses_result}.
        \end{tablenotes}
        \end{threeparttable}
	\end{table}
	
	\subsection{Conditional Generation of Small Molecules}
	\par
\begin{figure}[H]
		\centering
		\resizebox{0.9\linewidth}{0.1\linewidth}{\input{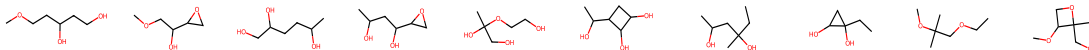}}
		\caption{Conditioned samples on QM9. The number of sampling steps was 1k. Since QM9 exhaustively included stable small molecules made up of CHONF, only 4 conditioned samples and 5 unconditioned samples are novel.}
		\label{fig:qm9_sample}
	\end{figure}

 The classifier-free guidance\cite{classifier-free} method is easily adapted into BFN, where only the computation of output distribution needs changing during sampling process. The pseudocode for computing discrete output distribution is presented in \prettyref{alg:classifier-free}. In the experiment, we jointly trained a model conditionally and unconditionally on QM9 dataset with an unconditional rate $p_{uncond}=0.2$. In the sampling stage, $w$ was set to 4. We sampled 10 molecules using the label [-0.249, 0.0615, 0.3105] that was transformed to $\boldsymbol{y}$ via a trained 2-layer MLP. 10 unconditioned samples were generated as a control group. RDKit\cite{rdkit} was employed to generate the 3D conformations, then the geometry optimisations and energy calculations were performed via PySCF\cite{pyscf} at B3LYP/6-31G(2df,p) level of accuracy. The results of MAE between calculated values and labels are presented in \prettyref{table:qm9_cond}. The conditioned samples are displayed in \prettyref{fig:qm9_sample}.
	\begin{algorithm}
		\caption{Invoking classifier-free guidance into output distribution}
		\label{alg:classifier-free}
		\begin{algorithmic}
			\Require $w\in\mathbb{R}$, conditioning vector $\boldsymbol{y}$
			\Function{DISCRETE\_OUTPUT\_DISTRIBUTION}{$\boldsymbol{\theta}\in [0,1]^{KD}$, $t\in[0,1]$, $\boldsymbol{y}\in\mathbb{R}^{f}$}
				\State Input ($\boldsymbol{\theta}$, $t$, $\boldsymbol{y}$) to network, receive $\boldsymbol{\Psi}(\boldsymbol{\theta},t,\boldsymbol{y})$ as output
			\If{in training stage or $\boldsymbol{y}$ is $\phi$}
				\State $\boldsymbol{p}_{O}(\cdot|\boldsymbol{\theta};t)\leftarrow {\rm softmax}(\boldsymbol{\Psi}(\boldsymbol{\theta},t,\boldsymbol{y}))_{dim=-1}$
			\Else
				\State Input ($\boldsymbol{\theta}$, $t$, $\phi$) to network, receive $\boldsymbol{\Psi}(\boldsymbol{\theta},t,\phi)$ as output
				\State $\boldsymbol{p}_{O}(\cdot|\boldsymbol{\theta};t)\leftarrow {\rm softmax}((1+w)\boldsymbol{\Psi}(\boldsymbol{\theta},t,\boldsymbol{y})-w\boldsymbol{\Psi}(\boldsymbol{\theta},t,\phi))_{dim=-1}$
			\EndIf
			\State {\bf return} $\boldsymbol{p}_{O}(\cdot|\boldsymbol{\theta};t)$
			\EndFunction
		\end{algorithmic}
	\end{algorithm}
	\begin{table}[H]
        \small
		\centering
		\sisetup{table-format=6.0}
		\setlength\cmidrulewidth{\heavyrulewidth}
		\caption{MAE on QM9 dataset w/ and w/o classifier-free guidance generation\textsuperscript{\emph{a}}}
        \label{table:qm9_cond}
        \begin{threeparttable}
		\begin{tabular}{lccc}
			\toprule
			{} & {$\epsilon_{HOMO}$ / a.u.} & {$\epsilon_{LUMO}$ / a.u.} & {$\Delta\epsilon$ / a.u.}\\
			\midrule
			{Conditional} & {\bf 0.00724} & {\bf 0.00981} & {\bf 0.01329}\\
			{Unconditional} & {0.01901} & {0.04076} & {0.04104}\\
			\bottomrule
		\end{tabular}
        \begin{tablenotes}
            \item [a] Smaller errors are in \textbf{\textit{bold}}.
        \end{tablenotes}
        \end{threeparttable}
	\end{table}
	
	\subsection{Molecular Scaffold Extension}
	\par
\begin{figure}[H]
		\centering
		\resizebox{0.9\linewidth}{0.15\linewidth}{\input{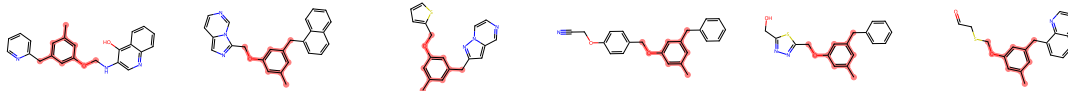}}
		\caption{An example of extended molecular scaffold. The scaffold is highlighted in red.}
		\label{fig:inpaint}
	\end{figure}

 Here, we show a simple \textit{inpaint} strategy can extend molecular scaffolds by using ChemBFN. In every sampling steps, parameters of input distributions are modified as $\boldsymbol{\theta}\leftarrow \boldsymbol{M}\odot\boldsymbol{e_{x}}+(1-\boldsymbol{M})\odot\boldsymbol{\theta}$ before being inputted into the network, where $\boldsymbol{M}$ is the mask and $\boldsymbol{e_{x}}$ is the one-hot representation of scaffold. \prettyref{fig:inpaint} shows an example of extending scaffold `Cc1cc(OC5)cc(C6)c1.' by a model trained on MOSES SAFE\cite{safe} version, a variation of SMILES. We found that inpainting sampling for 10 to 100 steps was sufficient to generate complex molecules.
	
	\subsection{Finetuning on Prediction Tasks}
	\par In this section, we compare our model with SOTA models\cite{uni-mol,molkd,gem,mole-bert,car,chemberta,chemberta2,smilestransformer,agbt}, including graph-based and language-based which could be further classified as smaller scale natural language processing models (NLPs) and large language models (LLMs), on subsets of MoleculeNet benchmark. As shown in \prettyref{table:moleculenet_result}, our method outperformed SOTA language models on several tasks, especially ClinTox and BBBP. It is notable that ChemBERTa\cite{chemberta} and ChemBERTa-2\cite{chemberta2}, which had a similar model size with ours, were pretrained on 77M molecules but had worse scores on 3 out of 5 tasks than ours. This indicated that BFN-style generative pretraining is a better strategy than masked language modeling and multitask regression pretraining. A similar observation applied to CaR$_{RoBERTa}$ model that coupled the knowledge of ChatGPT\cite{chatgpt} (which is far larger in scale than ours and is believed to have seen more chemical texts) and the distillation capability of RoBERTa\cite{roberta} method: our model outperformed CaR$_{RoBERTa}$ on 4 out of 5 tasks. However, when comparing with graph neural networks (GNNs) our model performed averagely 1.7\% worse, especially on regression tasks.
	\begin{table}[H]
		\centering
		\sisetup{table-format=6.0}
		\setlength\cmidrulewidth{\heavyrulewidth}
		\caption{Testing metrics on sub-tasks of MoleculeNet benchmark with {\bf scaffold splitting} compared with SOTA models\textsuperscript{\emph{a}}}
        \label{table:moleculenet_result}
        \Makebox[\textwidth][c]{
        \begin{threeparttable}
        \scriptsize
		\begin{tabular}{cl|cccc|ccc}
			\toprule
			\multicolumn{2}{c}{\multirow{2}*{\tabincell{c}{Method\\}}} & \multicolumn{4}{c}{ROC-AUC $\uparrow$} & \multicolumn{3}{c}{RMSE $\downarrow$}\\
			\multicolumn{2}{c}{} & {ClinTox} & {BBBP} & {BACE} & {HIV} & {ESOL} & {FreeSolv} & {Lipo}\\
			\midrule
			\midrule
			\multirow{4}*{\tabincell{c}{\begin{turn}{90}\tiny GNNs\end{turn}\\}} & {\tiny Uni-Mol\cite{uni-mol}} & {\underline{91.9} {\tiny $\pm$ 1.8}} & {72.9 {\tiny $\pm$ 0.6}} & {\underline{\bf 85.7} {\tiny $\pm$ 0.2}} & {\underline{\bf 80.8} {\tiny $\pm$ 0.3}} & {\underline{\bf 0.788} {\tiny $\pm$ 0.029}} & {\underline{1.480} {\tiny $\pm$ 0.048}} & {\underline{\bf 0.603} {\tiny $\pm$ 0.010}}\\
			{} & {\tiny MolKD\cite{molkd}} & {83.8 {\tiny $\pm$ 3.1}} & {\underline{74.8} {\tiny $\pm$ 2.3}} & {80.1 {\tiny $\pm$ 0.8}} & {74.9 {\tiny $\pm$ 1.7}} & {--} & {--} & {--}\\
			{} & {\tiny GEM\cite{gem}} & {90.1 {\tiny $\pm$ 1.3}} & {72.4 {\tiny $\pm$ 0.4}} & {85.6 {\tiny $\pm$ 1.1}} & {80.6 {\tiny $\pm$ 0.9}} & {0.798 {\tiny $\pm$ 0.029}} & {1.877 {\tiny $\pm$ 0.094}} & {0.660 {\tiny $\pm$ 0.008}}\\
			{} & {\tiny Mole-BERT\cite{mole-bert}} & {78.9 {\tiny $\pm$ 3.0}} & {71.9 {\tiny $\pm$ 1.6}} & {80.8 {\tiny $\pm$ 1.4}} & {78.2 {\tiny $\pm$ 0.8}} & {1.015 {\tiny $\pm$ 0.030}} & {--} & {0.676 {\tiny $\pm$ 0.017}}\\
			\midrule
			\midrule
			\multirow{2}*{\tabincell{c}{\begin{turn}{90}\tiny LLMs\end{turn}\\}} & \multirow{2}*{\tabincell{c}{\tiny CaR$_{RoBERTa}$\cite{car}\\}} & \multirow{2}*{\tabincell{c}{84.16 {\tiny $\pm$ 17.63}\\}} & \multirow{2}*{\tabincell{c}{81.99 {\tiny $\pm$ 4.19}\\}} & \multirow{2}*{\tabincell{c}{\underline{80.73} {\tiny $\pm$ 1.42}\\}} & \multirow{2}*{\tabincell{c}{--\\}} & \multirow{2}*{\tabincell{c}{0.96 {\tiny $\pm$ 0.09}\\}} & \multirow{2}*{\tabincell{c}{--\\}} & \multirow{2}*{\tabincell{c}{1.02 {\tiny $\pm$ 0.06}\\}}\\
			{} & {} & {} & {} & {} & {} & {} & {} & {}\\
			\midrule
			\multirow{7}*{\tabincell{c}{\begin{turn}{90}\tiny NLPs\end{turn}\\}} & {\tiny ChemBERTa\cite{chemberta}} & {73.3} & {64.3} & {--} & {62.2} & {--} & {--} & {--}\\
			{} & {\tiny ChemBERTa-2\cite{chemberta2}} & {60.1} & {74.2} & {79.9} & {--} & {--} & {--} & {\underline{0.744}}\\
                {} & {\tiny AGBT\cite{agbt}} & {--} & {76.3} & {--} & {--} & {--} & {--} & {--}\\
			{} & {\tiny SMILES Transformer\cite{smilestransformer}} & {--} & {70.4} & {70.1} & {72.9} & {--} & {--} & {--}\\
			{} & \cellcolor{gray!12}{\tiny {\bf ChemBFN} (ours)} & {\underline{\bf 99.18} {\tiny $\pm$ 1.77}} & {\underline{\bf 95.74} {\tiny $\pm$ 0.70}} & {73.56 {\tiny $\pm$ 1.22}} & {\underline{79.37} {\tiny $\pm$ 1.66}} & {\underline{0.884} {\tiny $\pm$ 0.003}} & {\underline{\bf 1.418} {\tiny $\pm$ 0.067}} & {0.746 {\tiny $\pm$ 0.001}}\\
			{} & \cellcolor{gray!12}{\tiny $\Delta_{GNNs_{best}}$} & \color{BrickRed}{+8\%} & \color{BrickRed}{+28\%} & \color{MidnightBlue}{-14\%} & \color{MidnightBlue}{-2\%} & \color{MidnightBlue}{+12\%} & \color{BrickRed}{-4\%} & \color{MidnightBlue}{+24\%}\\
			{} & \cellcolor{gray!12}{\tiny $\Delta_{LMs_{best}}$} & \color{BrickRed}{+18\%} & \color{BrickRed}{+17\%} & \color{MidnightBlue}{-9\%} & \color{BrickRed}{+9\%} & \color{BrickRed}{-8\%} & {--} & {0\%}\\
			\bottomrule
		\end{tabular}
        \small
        \begin{tablenotes}
            \item [a] The metrics of all other models were copied from their original paper. $\uparrow$ indicates that the higher is better and $\downarrow$ stands for the contrary. The best results are in \textbf{\textit{bold}}. The best results within the same category (graph-based or language-based) are \textit{\underline{underlined}}. Percentages in the last two rows show the performance changes w.r.t the best models and the colour represents whether our model was better (in {\color{BrickRed}red}) or not (in {\color{MidnightBlue}blue}).
        \end{tablenotes}
        \end{threeparttable}}
	\end{table}

    \par We further benchmarked our model on the public ADME dataset\cite{adme} (regression task) and Kinase inhibitor dataset\cite{kinase} (classification task). The results for the public ADME dataset were summarised in \prettyref{table:adme}. For the Kinase inhibitor dataset, the averaged ROC-AUC over 354 assays tested on the random split was (87.93 $\pm$ 14.05)\% and the averaged ROC-AUC tested on the scaffold split was (79.35 $\pm$ 18.71)\%.
    \begin{table}[H]
        \centering
        \sisetup{table-format=6.0}
		\setlength\cmidrulewidth{\heavyrulewidth}
		\caption{MAE, RMSE, and Pearson's correlation coefficient on the public ADME dataset.}
        \label{table:adme}
        \scriptsize
        \begin{tabular}{lcccccc}
            \toprule
            {} & {HLM} & {RLM} & {hPPB} & {rPPB} & {MDR1-MDCK ER} & {Solubility}\\
            \midrule
            \midrule
            {MAE} & {0.359 {\tiny $\pm$ 0.005}} & {0.428 {\tiny $\pm$ 0.001}} & {0.365 {\tiny $\pm$ 0.006}} & {0.408 {\tiny $\pm$ 0.006}} & {0.337 {\tiny $\pm$ 0.003}} & {0.411 {\tiny $\pm$ 0.008}}\\
            {RMSE} & {0.474 {\tiny $\pm$ 0.004}} & {0.556 {\tiny $\pm$ 0.005}} & {0.479 {\tiny $\pm$ 0.012}} & {0.549 {\tiny $\pm$ 0.009}} & {0.466 {\tiny $\pm$ 0.007}} & {0.630 {\tiny $\pm$ 0.013}}\\
            {R} & {0.653 {\tiny $\pm$ 0.002}} & {0.685 {\tiny $\pm$ 0.007}} & {0.771 {\tiny $\pm$ 0.001}} & {0.700 {\tiny $\pm$ 0.003}} & {0.765 {\tiny $\pm$ 0.007}} & {0.588 {\tiny $\pm$ 0.003}}\\
            \bottomrule
        \end{tabular}
    \end{table}
	
	\subsection{Reaction Yield Prediction}
	\par In order to predict the reaction yield, we first trained the generative model to understand chemical reaction by learning to predict the products. We developed an \textit{in-context} style guidance that during training stage, only the parameters of product in reaction SMILES were predicted. This was achieved by always masking the input distribution of reactant/reagent and \colorbox{gray!15}{$>>$} tokens that were converted to the corresponding one-hot representation, i.e., $\boldsymbol{\theta}\leftarrow \boldsymbol{M}_{rr}\odot\boldsymbol{e_{x}}+(1-\boldsymbol{M}_{rr})\odot\boldsymbol{\theta}$, where $\boldsymbol{M}_{rr}$ is the mask for reactant, reagent, and \colorbox{gray!15}{$>>$} token.
	\par The generative model was first pre-trained on USPTO-50k dataset then post-trained on Buchwald-Hartwig and Suzuki-Miyaura coupling datasets before the whole prediction model was fine-tuned. The testing scores compared with previous researches\cite{mff,yield-bert,yield-bert-da} were reported in \prettyref{table:reaction_yield}. It is notable that the Yield-BERT series\cite{yield-bert,yield-bert-da} were based on a pre-trained RXNFP\cite{rxnfp} model which had been pre-trained on over 2M reactions while our model was pre-trained on 50k reactions. Despite the disadvantage of limited access of pretraining data, the performance of our method was still close to that of largely pretrained model on random-split sets and significantly better on out-of-sample predictions.
	\begin{table}[H]
		\centering
		\sisetup{table-format=6.0}
		\setlength\cmidrulewidth{\heavyrulewidth}
		\caption{R$^{2}$ scores on different testing sets of HTE Buchwald-Hartwig and Suzuki-Miyaura reaction datasets\textsuperscript{\emph{a}}}
        \label{table:reaction_yield}
        \begin{threeparttable}
        \scriptsize
		\begin{tabular}{ll|cccc}
			\toprule
			\multicolumn{2}{c}{} & \multicolumn{4}{c}{Method}\\
			{Dataset} & {Split} & {MFF\cite{mff}} & {Yield-BERT\cite{yield-bert}} & {Yield-BERT-DA\cite{yield-bert-da}} & {ChemBFN (ours)}\\
			\midrule
			\midrule
			\multirow{6}*{\tabincell{l}{Buchwald-\\Hartwig\\}} & {Rand 70/30} & {0.927 $\pm$ {\tiny 0.007}} & {0.951 $\pm$ {\tiny 0.005}} & {{\bf 0.969} $\pm$ {\tiny 0.004}} & {0.952 $\pm$ {\tiny 0.008}}\\
			{} & {Test 1} & {\bf 0.85} & {0.84 {\tiny $\pm$ 0.010}} & {0.82 {\tiny $\pm$ 0.01}} & {0.844 {\tiny $\pm$ 0.002}}\\
			{} & {Test 2} & {0.71} & {0.84 {\tiny $\pm$ 0.03}} & {0.90 {\tiny $\pm$ 0.01}} & {{\bf 0.910} {\tiny $\pm$ 0.001}}\\
			{} & {Test 3} & {0.64} & {0.75 {\tiny $\pm$ 0.04}} & {0.63 {\tiny $\pm$ 0.05}} & {{\bf 0.787} {\tiny $\pm$ 0.034}}\\
			{} & {Test 4} & {0.18} & {0.49 {\tiny $\pm$ 0.05}} & {0.43 {\tiny $\pm$ 0.07}} & {{\bf 0.633}  {\tiny $\pm$ 0.082}}\\
			{} & {Avg. 1-4} & {0.60} & {0.73 {\tiny $\pm$ 0.15}} & {0.69 {\tiny $\pm$ 0.19}} & {{\bf 0.794} {\tiny $\pm$ 0.118}}\\
			\midrule
			\multirow{2}*{\tabincell{l}{Suzuki-\\Miyaura}} & \multirow{2}*{\tabincell{l}{Rand 70/30\\}} & \multirow{2}*{\tabincell{l}{--\\}} & \multirow{2}*{\tabincell{l}{{\bf 0.81} $\pm$ {\tiny 0.02}\\}} & \multirow{2}*{\tabincell{l}{--\\}} & \multirow{2}*{\tabincell{l}{0.796 $\pm$ {\tiny 0.011}\\}}\\
			{} & {} & {} & {} & {} & {}\\
			\bottomrule
		\end{tabular}
        \small
        \begin{tablenotes}
            \item [a] The scores of all other models were copied from the original paper. The best results are in \textbf{\textit{bold}}. The score of ``rand 70/30'' split was the 10-fold average value. Test 1-4 were out-of-sample splits.
        \end{tablenotes}
        \end{threeparttable}
	\end{table}
	
	\subsection{Is Larger Pretrain Dataset Better?}
	\par We have seen that our model, although was pretrained on 40M molecules, outperformed the models pretrained on larger dataset on several prediction tasks. Here rises a question: does a larger pretraining dataset benefit our method? To answer this, three models were trained on AqSolDB dataset, of which one was trained from scratch, one was pretrained on 40M molecules from ZINC15 database, and the third one was pretrained on 190M molecules from ZINC15. We summarised the testing results in \prettyref{table:size}. Interestingly, the errors did not shrink when the pretraining data grew from 40M to 190M. However, compared with zero pretraining, an improvement in performance of $\geq$12.5\% can be confirmed.
	\begin{table}[H]
        \small
		\centering
		\sisetup{table-format=6.0}
		\setlength\cmidrulewidth{\heavyrulewidth}
		\caption{Testing metrics of models with different pretrain data sizes (0, 40M, and 190M) on AqSolDB dataset}
        \label{table:size}
		\begin{tabular}{lccc}
			\toprule
			{} & {From scratch} & {Pretrained on 40M} & {Pretrained on 190M}\\
			\midrule
			\midrule
			{MAE} & {0.978 {\scriptsize $\pm$ 0.016}} & {0.837 {\scriptsize $\pm$ 0.005}} & {0.851 {\scriptsize $\pm$ 0.021}}\\
			{RMSE} & {1.309 {\scriptsize $\pm$ 0.014}} & {1.131 {\scriptsize $\pm$ 0.008}} & {1.145 {\scriptsize $\pm$ 0.034}}\\
			\bottomrule
		\end{tabular}
	\end{table}
	
	\subsection{Training Details}
	\par For all generative tasks, the models were trained for 100 epochs with the batch-size of 120 molecule/batch. The learning rate ($lr$) was $5.0\times 10^{-5}$ that was linearly increased (warm-up) from $10^{-8}$ during the first 1,000 training steps.
	\par We pre-trained one model on 40M SMILES for 15 epochs with the batch-size of 512 on single A100 GPU and one model on 190M SMILES for 5 epochs with the effective batch-size of 1,024 ($2\times 512$) on 2$\times$A100 GPUs. The warm-up strategy and $lr$ were the same as mentioned above.
	\par During fine-tuning stages, models were trained for 100 epochs on labelled datasets. The batch-size, both for training and validation, was 32 on MoleculeNet benchmark, AqSolDB dataset, public ADME dataset, and Kinase inhibitor dataset; the training batch-size was 16 for reaction yield prediction. $lr_{max}$ was $10^{-4}$ that was warmed up from $10^{-7}$ during the first 1,000 steps for regression tasks and 100 steps for classification tasks. After warm-up stage, $lr$ decreased by 0.2 after the validation metrics stopped improving for 20 epochs unless the learning rate had reached $10^{-6}$. The dropout rate of prediction MLP head was fine-tuned for each case and we recommend to try from $\{0.0,0.1,0.5,0.7\}$. The validation metrics for regression and classification tasks were MAE and inverted accuracy (i.e., 1 - accuracy), respectively.
	\par We employed AdamW\cite{adamw} with default hyperparameters implemented in PyTorch\cite{pytorch} as the optimizer for all tasks.
	
	\section{Conclusion}
	\par ChemBFN, a Bayesian flow network framework for chemistry tasks of both generation and prediction, was developed in this work. The new accuracy schedule helped ChemBFN achieve competitive performance of discrete diffusion models and autoregressive models on generating large molecules. We proposed a BFN-style generative pretraining strategy that surpassed existing language-based transformer models on several classification and regression tasks. We believe this work provides a tool that can accelerate researches of both drug designing and filtering and give in helpful information for synthesis planning. However, we still leave gaps between graph-based models in prediction tasks, which we shall keep for the future research.

    \section{Data and Software Availability}
	\par The code, pre-trained models, and instructions necessary to reproduce the results of this study are available for download at https://github.com/Augus1999/bayesian-flow-network-for-chemistry.
	
	\section{Acknowledgements}
	\par We express our gratitude to the Research Center for Computational Science (RCCS) in Okazaki, Japan, and its maintenance team for providing computing resources, including A100 GPUs. This work is under project 24-IMS-C043 of RCCS. We also thank Dr. Maho Nakata who kindly lent us his own RTX 3080 GPU and Prof. Kazumasa Okada for discussion.

    \section{Conflict of Interest}
    \par There is no conflict of interest.

    \section{Funding Sources}
    \par The authors claim that there is no funding related to this research.

	\bibliography{chembfn.bib}

\providecommand{\latin}[1]{#1}
\makeatletter
\providecommand{\doi}
  {\begingroup\let\do\@makeother\dospecials
  \catcode`\{=1 \catcode`\}=2 \doi@aux}
\providecommand{\doi@aux}[1]{\endgroup\texttt{#1}}
\makeatother
\providecommand*\mcitethebibliography{\thebibliography}
\csname @ifundefined\endcsname{endmcitethebibliography}
  {\let\endmcitethebibliography\endthebibliography}{}
\begin{mcitethebibliography}{62}
\providecommand*\natexlab[1]{#1}
\providecommand*\mciteSetBstSublistMode[1]{}
\providecommand*\mciteSetBstMaxWidthForm[2]{}
\providecommand*\mciteBstWouldAddEndPuncttrue
  {\def\EndOfBibitem{\unskip.}}
\providecommand*\mciteBstWouldAddEndPunctfalse
  {\let\EndOfBibitem\relax}
\providecommand*\mciteSetBstMidEndSepPunct[3]{}
\providecommand*\mciteSetBstSublistLabelBeginEnd[3]{}
\providecommand*\EndOfBibitem{}
\mciteSetBstSublistMode{f}
\mciteSetBstMaxWidthForm{subitem}{(\alph{mcitesubitemcount})}
\mciteSetBstSublistLabelBeginEnd
  {\mcitemaxwidthsubitemform\space}
  {\relax}
  {\relax}

\bibitem[Segler \latin{et~al.}(2018)Segler, Kogej, Tyrchan, and
  Waller]{generating}
Segler,~M.~H.; Kogej,~T.; Tyrchan,~C.; Waller,~M.~P. Generating focused
  molecule libraries for drug discovery with recurrent neural networks.
  \emph{ACS central science} \textbf{2018}, \emph{4}, 120--131\relax
\mciteBstWouldAddEndPuncttrue
\mciteSetBstMidEndSepPunct{\mcitedefaultmidpunct}
{\mcitedefaultendpunct}{\mcitedefaultseppunct}\relax
\EndOfBibitem
\bibitem[Amabilino \latin{et~al.}(2020)Amabilino, Pog{\'a}ny, Pickett, and
  Green]{rnn-guidelines}
Amabilino,~S.; Pog{\'a}ny,~P.; Pickett,~S.~D.; Green,~D.~V. Guidelines for
  recurrent neural network transfer learning-based molecular generation of
  focused libraries. \emph{Journal of Chemical Information and Modeling}
  \textbf{2020}, \emph{60}, 5699--5713\relax
\mciteBstWouldAddEndPuncttrue
\mciteSetBstMidEndSepPunct{\mcitedefaultmidpunct}
{\mcitedefaultendpunct}{\mcitedefaultseppunct}\relax
\EndOfBibitem
\bibitem[Prykhodko \latin{et~al.}(2019)Prykhodko, Johansson, Kotsias,
  Ar{\'u}s-Pous, Bjerrum, Engkvist, and Chen]{latentgan}
Prykhodko,~O.; Johansson,~S.~V.; Kotsias,~P.-C.; Ar{\'u}s-Pous,~J.;
  Bjerrum,~E.~J.; Engkvist,~O.; Chen,~H. A de novo molecular generation method
  using latent vector based generative adversarial network. \emph{Journal of
  Cheminformatics} \textbf{2019}, \emph{11}, 74\relax
\mciteBstWouldAddEndPuncttrue
\mciteSetBstMidEndSepPunct{\mcitedefaultmidpunct}
{\mcitedefaultendpunct}{\mcitedefaultseppunct}\relax
\EndOfBibitem
\bibitem[Bagal \latin{et~al.}(2021)Bagal, Aggarwal, Vinod, and
  Priyakumar]{molgpt}
Bagal,~V.; Aggarwal,~R.; Vinod,~P.; Priyakumar,~U.~D. MolGPT: molecular
  generation using a transformer-decoder model. \emph{Journal of Chemical
  Information and Modeling} \textbf{2021}, \emph{62}, 2064--2076\relax
\mciteBstWouldAddEndPuncttrue
\mciteSetBstMidEndSepPunct{\mcitedefaultmidpunct}
{\mcitedefaultendpunct}{\mcitedefaultseppunct}\relax
\EndOfBibitem
\bibitem[Jin \latin{et~al.}(2018)Jin, Barzilay, and Jaakkola]{jtn-vae}
Jin,~W.; Barzilay,~R.; Jaakkola,~T. Junction tree variational autoencoder for
  molecular graph generation. International conference on machine learning.
  2018; pp 2323--2332\relax
\mciteBstWouldAddEndPuncttrue
\mciteSetBstMidEndSepPunct{\mcitedefaultmidpunct}
{\mcitedefaultendpunct}{\mcitedefaultseppunct}\relax
\EndOfBibitem
\bibitem[Brown \latin{et~al.}(2019)Brown, Fiscato, Segler, and
  Vaucher]{guacamol}
Brown,~N.; Fiscato,~M.; Segler,~M.~H.; Vaucher,~A.~C. GuacaMol: benchmarking
  models for de novo molecular design. \emph{Journal of chemical information
  and modeling} \textbf{2019}, \emph{59}, 1096--1108\relax
\mciteBstWouldAddEndPuncttrue
\mciteSetBstMidEndSepPunct{\mcitedefaultmidpunct}
{\mcitedefaultendpunct}{\mcitedefaultseppunct}\relax
\EndOfBibitem
\bibitem[Loeffler \latin{et~al.}(2024)Loeffler, He, Tibo, Janet, Voronov,
  Mervin, and Engkvist]{reinvent4}
Loeffler,~H.~H.; He,~J.; Tibo,~A.; Janet,~J.~P.; Voronov,~A.; Mervin,~L.~H.;
  Engkvist,~O. Reinvent 4: Modern AI--driven generative molecule design.
  \emph{Journal of Cheminformatics} \textbf{2024}, \emph{16}, 20\relax
\mciteBstWouldAddEndPuncttrue
\mciteSetBstMidEndSepPunct{\mcitedefaultmidpunct}
{\mcitedefaultendpunct}{\mcitedefaultseppunct}\relax
\EndOfBibitem
\bibitem[Guo \latin{et~al.}(2023)Guo, Knuth, Margreitter, Janet, Papadopoulos,
  Engkvist, and Patronov]{link-invent}
Guo,~J.; Knuth,~F.; Margreitter,~C.; Janet,~J.~P.; Papadopoulos,~K.;
  Engkvist,~O.; Patronov,~A. Link-INVENT: generative linker design with
  reinforcement learning. \emph{Digital Discovery} \textbf{2023}, \emph{2},
  392--408\relax
\mciteBstWouldAddEndPuncttrue
\mciteSetBstMidEndSepPunct{\mcitedefaultmidpunct}
{\mcitedefaultendpunct}{\mcitedefaultseppunct}\relax
\EndOfBibitem
\bibitem[Popova \latin{et~al.}(2018)Popova, Isayev, and Tropsha]{release}
Popova,~M.; Isayev,~O.; Tropsha,~A. Deep reinforcement learning for de novo
  drug design. \emph{Science advances} \textbf{2018}, \emph{4}, eaap7885\relax
\mciteBstWouldAddEndPuncttrue
\mciteSetBstMidEndSepPunct{\mcitedefaultmidpunct}
{\mcitedefaultendpunct}{\mcitedefaultseppunct}\relax
\EndOfBibitem
\bibitem[Mercado \latin{et~al.}(2021)Mercado, Rastemo, Lindelöf, Klambauer,
  Engkvist, Chen, and Bjerrum]{graphinvent}
Mercado,~R.; Rastemo,~T.; Lindelöf,~E.; Klambauer,~G.; Engkvist,~O.; Chen,~H.;
  Bjerrum,~E.~J. Graph networks for molecular design. \emph{Machine Learning:
  Science and Technology} \textbf{2021}, \emph{2}, 025023\relax
\mciteBstWouldAddEndPuncttrue
\mciteSetBstMidEndSepPunct{\mcitedefaultmidpunct}
{\mcitedefaultendpunct}{\mcitedefaultseppunct}\relax
\EndOfBibitem
\bibitem[Jensen(2019)]{mcts}
Jensen,~J.~H. A graph-based genetic algorithm and generative model/Monte Carlo
  tree search for the exploration of chemical space. \emph{Chemical science}
  \textbf{2019}, \emph{10}, 3567--3572\relax
\mciteBstWouldAddEndPuncttrue
\mciteSetBstMidEndSepPunct{\mcitedefaultmidpunct}
{\mcitedefaultendpunct}{\mcitedefaultseppunct}\relax
\EndOfBibitem
\bibitem[Iwata \latin{et~al.}(2023)Iwata, Nakai, Koyama, Matsumoto, Kojima, and
  Okuno]{vgae-mcts}
Iwata,~H.; Nakai,~T.; Koyama,~T.; Matsumoto,~S.; Kojima,~R.; Okuno,~Y.
  VGAE-MCTS: A New Molecular Generative Model Combining the Variational Graph
  Auto-Encoder and Monte Carlo Tree Search. \emph{Journal of Chemical
  Information and Modeling} \textbf{2023}, \emph{63}, 7392--7400\relax
\mciteBstWouldAddEndPuncttrue
\mciteSetBstMidEndSepPunct{\mcitedefaultmidpunct}
{\mcitedefaultendpunct}{\mcitedefaultseppunct}\relax
\EndOfBibitem
\bibitem[Yang \latin{et~al.}(2017)Yang, Zhang, Yoshizoe, Terayama, and
  Tsuda]{chemts}
Yang,~X.; Zhang,~J.; Yoshizoe,~K.; Terayama,~K.; Tsuda,~K. ChemTS: an efficient
  python library for de novo molecular generation. \emph{Science and technology
  of advanced materials} \textbf{2017}, \emph{18}, 972--976\relax
\mciteBstWouldAddEndPuncttrue
\mciteSetBstMidEndSepPunct{\mcitedefaultmidpunct}
{\mcitedefaultendpunct}{\mcitedefaultseppunct}\relax
\EndOfBibitem
\bibitem[Li \latin{et~al.}(2018)Li, Zhang, and Liu]{molrnn}
Li,~Y.; Zhang,~L.; Liu,~Z. Multi-objective de novo drug design with conditional
  graph generative model. \emph{Journal of cheminformatics} \textbf{2018},
  \emph{10}, 33\relax
\mciteBstWouldAddEndPuncttrue
\mciteSetBstMidEndSepPunct{\mcitedefaultmidpunct}
{\mcitedefaultendpunct}{\mcitedefaultseppunct}\relax
\EndOfBibitem
\bibitem[Atance \latin{et~al.}(2022)Atance, Diez, Engkvist, Olsson, and
  Mercado]{dgm}
Atance,~S.~R.; Diez,~J.~V.; Engkvist,~O.; Olsson,~S.; Mercado,~R. De novo drug
  design using reinforcement learning with graph-based deep generative models.
  \emph{Journal of chemical information and modeling} \textbf{2022}, \emph{62},
  4863--4872\relax
\mciteBstWouldAddEndPuncttrue
\mciteSetBstMidEndSepPunct{\mcitedefaultmidpunct}
{\mcitedefaultendpunct}{\mcitedefaultseppunct}\relax
\EndOfBibitem
\bibitem[Polykovskiy \latin{et~al.}(2020)Polykovskiy, Zhebrak,
  Sanchez-Lengeling, Golovanov, Tatanov, Belyaev, Kurbanov, Artamonov,
  Aladinskiy, Veselov, Kadurin, Johansson, Chen, Nikolenko, Aspuru-Guzik, and
  Zhavoronkov]{moses}
Polykovskiy,~D. \latin{et~al.}  Molecular Sets (MOSES): A Benchmarking Platform
  for Molecular Generation Models. 2020;
  \url{https://arxiv.org/abs/1811.12823}\relax
\mciteBstWouldAddEndPuncttrue
\mciteSetBstMidEndSepPunct{\mcitedefaultmidpunct}
{\mcitedefaultendpunct}{\mcitedefaultseppunct}\relax
\EndOfBibitem
\bibitem[Ho \latin{et~al.}(2020)Ho, Jain, and Abbeel]{ddpm}
Ho,~J.; Jain,~A.; Abbeel,~P. Denoising diffusion probabilistic models.
  \emph{Advances in neural information processing systems} \textbf{2020},
  \emph{33}, 6840--6851\relax
\mciteBstWouldAddEndPuncttrue
\mciteSetBstMidEndSepPunct{\mcitedefaultmidpunct}
{\mcitedefaultendpunct}{\mcitedefaultseppunct}\relax
\EndOfBibitem
\bibitem[Vignac \latin{et~al.}(2023)Vignac, Krawczuk, Siraudin, Wang, Cevher,
  and Frossard]{digress}
Vignac,~C.; Krawczuk,~I.; Siraudin,~A.; Wang,~B.; Cevher,~V.; Frossard,~P.
  DiGress: Discrete Denoising diffusion for graph generation. 2023;
  \url{https://arxiv.org/abs/2209.14734}\relax
\mciteBstWouldAddEndPuncttrue
\mciteSetBstMidEndSepPunct{\mcitedefaultmidpunct}
{\mcitedefaultendpunct}{\mcitedefaultseppunct}\relax
\EndOfBibitem
\bibitem[Graves \latin{et~al.}(2024)Graves, Srivastava, Atkinson, and
  Gomez]{bfn}
Graves,~A.; Srivastava,~R.~K.; Atkinson,~T.; Gomez,~F. Bayesian Flow Networks.
  2024; \url{https://arxiv.org/abs/2308.07037}\relax
\mciteBstWouldAddEndPuncttrue
\mciteSetBstMidEndSepPunct{\mcitedefaultmidpunct}
{\mcitedefaultendpunct}{\mcitedefaultseppunct}\relax
\EndOfBibitem
\bibitem[Song \latin{et~al.}(2024)Song, Gong, Zhou, Zheng, Liu, and Ma]{geobfn}
Song,~Y.; Gong,~J.; Zhou,~H.; Zheng,~M.; Liu,~J.; Ma,~W.-Y. Unified Generative
  Modeling of 3D Molecules with Bayesian Flow Networks. The Twelfth
  International Conference on Learning Representations. 2024\relax
\mciteBstWouldAddEndPuncttrue
\mciteSetBstMidEndSepPunct{\mcitedefaultmidpunct}
{\mcitedefaultendpunct}{\mcitedefaultseppunct}\relax
\EndOfBibitem
\bibitem[Weininger(1988)]{smiles}
Weininger,~D. SMILES, a chemical language and information system. 1.
  Introduction to methodology and encoding rules. \emph{Journal of chemical
  information and computer sciences} \textbf{1988}, \emph{28}, 31--36\relax
\mciteBstWouldAddEndPuncttrue
\mciteSetBstMidEndSepPunct{\mcitedefaultmidpunct}
{\mcitedefaultendpunct}{\mcitedefaultseppunct}\relax
\EndOfBibitem
\bibitem[Krenn \latin{et~al.}(2020)Krenn, H{\"a}se, Nigam, Friederich, and
  Aspuru-Guzik]{selfies}
Krenn,~M.; H{\"a}se,~F.; Nigam,~A.; Friederich,~P.; Aspuru-Guzik,~A.
  Self-referencing embedded strings (SELFIES): A 100\% robust molecular string
  representation. \emph{Machine Learning: Science and Technology}
  \textbf{2020}, \emph{1}, 045024\relax
\mciteBstWouldAddEndPuncttrue
\mciteSetBstMidEndSepPunct{\mcitedefaultmidpunct}
{\mcitedefaultendpunct}{\mcitedefaultseppunct}\relax
\EndOfBibitem
\bibitem[Vaswani \latin{et~al.}(2017)Vaswani, Shazeer, Parmar, Uszkoreit,
  Jones, Gomez, Kaiser, and Polosukhin]{attention}
Vaswani,~A.; Shazeer,~N.; Parmar,~N.; Uszkoreit,~J.; Jones,~L.; Gomez,~A.~N.;
  Kaiser,~L.~u.; Polosukhin,~I. Attention is All you Need. Advances in Neural
  Information Processing Systems. 2017\relax
\mciteBstWouldAddEndPuncttrue
\mciteSetBstMidEndSepPunct{\mcitedefaultmidpunct}
{\mcitedefaultendpunct}{\mcitedefaultseppunct}\relax
\EndOfBibitem
\bibitem[Peebles and Xie(2023)Peebles, and Xie]{dit}
Peebles,~W.; Xie,~S. Scalable diffusion models with transformers. Proceedings
  of the IEEE/CVF International Conference on Computer Vision. 2023; pp
  4195--4205\relax
\mciteBstWouldAddEndPuncttrue
\mciteSetBstMidEndSepPunct{\mcitedefaultmidpunct}
{\mcitedefaultendpunct}{\mcitedefaultseppunct}\relax
\EndOfBibitem
\bibitem[Klambauer \latin{et~al.}(2017)Klambauer, Unterthiner, Mayr, and
  Hochreiter]{selu}
Klambauer,~G.; Unterthiner,~T.; Mayr,~A.; Hochreiter,~S. Self-Normalizing
  Neural Networks. Advances in Neural Information Processing Systems.
  2017\relax
\mciteBstWouldAddEndPuncttrue
\mciteSetBstMidEndSepPunct{\mcitedefaultmidpunct}
{\mcitedefaultendpunct}{\mcitedefaultseppunct}\relax
\EndOfBibitem
\bibitem[Sun \latin{et~al.}(2022)Sun, Dong, Patra, Ma, Huang, Benhaim,
  Chaudhary, Song, and Wei]{xpos}
Sun,~Y.; Dong,~L.; Patra,~B.; Ma,~S.; Huang,~S.; Benhaim,~A.; Chaudhary,~V.;
  Song,~X.; Wei,~F. A Length-Extrapolatable Transformer. 2022;
  \url{https://arxiv.org/abs/2212.10554}\relax
\mciteBstWouldAddEndPuncttrue
\mciteSetBstMidEndSepPunct{\mcitedefaultmidpunct}
{\mcitedefaultendpunct}{\mcitedefaultseppunct}\relax
\EndOfBibitem
\bibitem[Su \latin{et~al.}(2024)Su, Ahmed, Lu, Pan, Bo, and Liu]{roformer}
Su,~J.; Ahmed,~M.; Lu,~Y.; Pan,~S.; Bo,~W.; Liu,~Y. RoFormer: Enhanced
  transformer with Rotary Position Embedding. \emph{Neurocomputing}
  \textbf{2024}, \emph{568}, 127063\relax
\mciteBstWouldAddEndPuncttrue
\mciteSetBstMidEndSepPunct{\mcitedefaultmidpunct}
{\mcitedefaultendpunct}{\mcitedefaultseppunct}\relax
\EndOfBibitem
\bibitem[Devlin \latin{et~al.}(2019)Devlin, Chang, Lee, and Toutanova]{bert}
Devlin,~J.; Chang,~M.-W.; Lee,~K.; Toutanova,~K. BERT: Pre-training of Deep
  Bidirectional Transformers for Language Understanding. 2019;
  \url{https://arxiv.org/abs/1810.04805}\relax
\mciteBstWouldAddEndPuncttrue
\mciteSetBstMidEndSepPunct{\mcitedefaultmidpunct}
{\mcitedefaultendpunct}{\mcitedefaultseppunct}\relax
\EndOfBibitem
\bibitem[Liu \latin{et~al.}(2019)Liu, Ott, Goyal, Du, Joshi, Chen, Levy, Lewis,
  Zettlemoyer, and Stoyanov]{roberta}
Liu,~Y.; Ott,~M.; Goyal,~N.; Du,~J.; Joshi,~M.; Chen,~D.; Levy,~O.; Lewis,~M.;
  Zettlemoyer,~L.; Stoyanov,~V. RoBERTa: A Robustly Optimized BERT Pretraining
  Approach. 2019; \url{https://arxiv.org/abs/1907.11692}\relax
\mciteBstWouldAddEndPuncttrue
\mciteSetBstMidEndSepPunct{\mcitedefaultmidpunct}
{\mcitedefaultendpunct}{\mcitedefaultseppunct}\relax
\EndOfBibitem
\bibitem[Zhang \latin{et~al.}(2022)Zhang, Zhang, Bao, and
  Wei]{attention-temperature}
Zhang,~S.; Zhang,~X.; Bao,~H.; Wei,~F. Attention Temperature Matters in
  Abstractive Summarization Distillation. ACL 2022. 2022\relax
\mciteBstWouldAddEndPuncttrue
\mciteSetBstMidEndSepPunct{\mcitedefaultmidpunct}
{\mcitedefaultendpunct}{\mcitedefaultseppunct}\relax
\EndOfBibitem
\bibitem[Preuer \latin{et~al.}(2018)Preuer, Renz, Unterthiner, Hochreiter, and
  Klambauer]{fcd}
Preuer,~K.; Renz,~P.; Unterthiner,~T.; Hochreiter,~S.; Klambauer,~G. Fréchet
  ChemNet Distance: A Metric for Generative Models for Molecules in Drug
  Discovery. \emph{Journal of Chemical Information and Modeling} \textbf{2018},
  \emph{58}, 1736--1741, PMID: 30118593\relax
\mciteBstWouldAddEndPuncttrue
\mciteSetBstMidEndSepPunct{\mcitedefaultmidpunct}
{\mcitedefaultendpunct}{\mcitedefaultseppunct}\relax
\EndOfBibitem
\bibitem[Degen \latin{et~al.}(2008)Degen, Wegscheid-Gerlach, Zaliani, and
  Rarey]{brics}
Degen,~J.; Wegscheid-Gerlach,~C.; Zaliani,~A.; Rarey,~M. On the Art of
  Compiling and Using 'Drug-Like' Chemical Fragment Spaces. \emph{ChemMedChem}
  \textbf{2008}, \emph{3}, 1503--1507\relax
\mciteBstWouldAddEndPuncttrue
\mciteSetBstMidEndSepPunct{\mcitedefaultmidpunct}
{\mcitedefaultendpunct}{\mcitedefaultseppunct}\relax
\EndOfBibitem
\bibitem[Bemis and Murcko(1996)Bemis, and Murcko]{bemis-murcko-scaffold}
Bemis,~G.~W.; Murcko,~M.~A. The Properties of Known Drugs. 1. Molecular
  Frameworks. \emph{Journal of Medicinal Chemistry} \textbf{1996}, \emph{39},
  2887--2893, PMID: 8709122\relax
\mciteBstWouldAddEndPuncttrue
\mciteSetBstMidEndSepPunct{\mcitedefaultmidpunct}
{\mcitedefaultendpunct}{\mcitedefaultseppunct}\relax
\EndOfBibitem
\bibitem[Ramakrishnan \latin{et~al.}(2014)Ramakrishnan, Dral, Rupp, and
  Von~Lilienfeld]{qm9}
Ramakrishnan,~R.; Dral,~P.~O.; Rupp,~M.; Von~Lilienfeld,~O.~A. Quantum
  chemistry structures and properties of 134 kilo molecules. \emph{Scientific
  data} \textbf{2014}, \emph{1}, 140022\relax
\mciteBstWouldAddEndPuncttrue
\mciteSetBstMidEndSepPunct{\mcitedefaultmidpunct}
{\mcitedefaultendpunct}{\mcitedefaultseppunct}\relax
\EndOfBibitem
\bibitem[Sterling and Irwin(2015)Sterling, and Irwin]{zinc15}
Sterling,~T.; Irwin,~J.~J. ZINC 15--ligand discovery for everyone.
  \emph{Journal of chemical information and modeling} \textbf{2015}, \emph{55},
  2324--2337\relax
\mciteBstWouldAddEndPuncttrue
\mciteSetBstMidEndSepPunct{\mcitedefaultmidpunct}
{\mcitedefaultendpunct}{\mcitedefaultseppunct}\relax
\EndOfBibitem
\bibitem[Wu \latin{et~al.}(2018)Wu, Ramsundar, Feinberg, Gomes, Geniesse,
  Pappu, Leswing, and Pande]{moleculenet}
Wu,~Z.; Ramsundar,~B.; Feinberg,~E.~N.; Gomes,~J.; Geniesse,~C.; Pappu,~A.~S.;
  Leswing,~K.; Pande,~V. MoleculeNet: a benchmark for molecular machine
  learning. \emph{Chemical science} \textbf{2018}, \emph{9}, 513--530\relax
\mciteBstWouldAddEndPuncttrue
\mciteSetBstMidEndSepPunct{\mcitedefaultmidpunct}
{\mcitedefaultendpunct}{\mcitedefaultseppunct}\relax
\EndOfBibitem
\bibitem[Ramsundar \latin{et~al.}(2019)Ramsundar, Eastman, Walters, Pande,
  Leswing, and Wu]{deepchem}
Ramsundar,~B.; Eastman,~P.; Walters,~P.; Pande,~V.; Leswing,~K.; Wu,~Z.
  \emph{Deep Learning for the Life Sciences}; O'Reilly Media, 2019;
  \url{https://www.amazon.com/Deep-Learning-Life-Sciences-Microscopy/dp/1492039837}\relax
\mciteBstWouldAddEndPuncttrue
\mciteSetBstMidEndSepPunct{\mcitedefaultmidpunct}
{\mcitedefaultendpunct}{\mcitedefaultseppunct}\relax
\EndOfBibitem
\bibitem[Fang \latin{et~al.}(2023)Fang, Wang, Grater, Kapadnis, Black, Trapa,
  and Sciabola]{adme}
Fang,~C.; Wang,~Y.; Grater,~R.; Kapadnis,~S.; Black,~C.; Trapa,~P.;
  Sciabola,~S. Prospective validation of machine learning algorithms for
  absorption, distribution, metabolism, and excretion prediction: An industrial
  perspective. \emph{Journal of Chemical Information and Modeling}
  \textbf{2023}, \emph{63}, 3263--3274\relax
\mciteBstWouldAddEndPuncttrue
\mciteSetBstMidEndSepPunct{\mcitedefaultmidpunct}
{\mcitedefaultendpunct}{\mcitedefaultseppunct}\relax
\EndOfBibitem
\bibitem[Wu \latin{et~al.}(2024)Wu, Chen, Wu, Zhao, Huang, Lin, and
  Wang]{kinase}
Wu,~J.; Chen,~Y.; Wu,~J.; Zhao,~D.; Huang,~J.; Lin,~M.; Wang,~L. Large-scale
  comparison of machine learning methods for profiling prediction of kinase
  inhibitors. \emph{Journal of Cheminformatics} \textbf{2024}, \emph{16},
  13\relax
\mciteBstWouldAddEndPuncttrue
\mciteSetBstMidEndSepPunct{\mcitedefaultmidpunct}
{\mcitedefaultendpunct}{\mcitedefaultseppunct}\relax
\EndOfBibitem
\bibitem[Schneider \latin{et~al.}(2016)Schneider, Stiefl, and
  Landrum]{uspto-50k}
Schneider,~N.; Stiefl,~N.; Landrum,~G.~A. What’s What: The (Nearly)
  Definitive Guide to Reaction Role Assignment. \emph{Journal of Chemical
  Information and Modeling} \textbf{2016}, \emph{56}, 2336--2346, PMID:
  28024398\relax
\mciteBstWouldAddEndPuncttrue
\mciteSetBstMidEndSepPunct{\mcitedefaultmidpunct}
{\mcitedefaultendpunct}{\mcitedefaultseppunct}\relax
\EndOfBibitem
\bibitem[Schwaller \latin{et~al.}(2021)Schwaller, Vaucher, Laino, and
  Reymond]{yield-bert}
Schwaller,~P.; Vaucher,~A.~C.; Laino,~T.; Reymond,~J.-L. Prediction of chemical
  reaction yields using deep learning. \emph{Machine Learning: Science and
  Technology} \textbf{2021}, \emph{2}, 015016\relax
\mciteBstWouldAddEndPuncttrue
\mciteSetBstMidEndSepPunct{\mcitedefaultmidpunct}
{\mcitedefaultendpunct}{\mcitedefaultseppunct}\relax
\EndOfBibitem
\bibitem[Sorkun \latin{et~al.}(2019)Sorkun, Khetan, and Er]{aqsoldb}
Sorkun,~M.~C.; Khetan,~A.; Er,~S. AqSolDB, a curated reference set of aqueous
  solubility and 2D descriptors for a diverse set of compounds.
  \emph{Scientific data} \textbf{2019}, \emph{6}, 143\relax
\mciteBstWouldAddEndPuncttrue
\mciteSetBstMidEndSepPunct{\mcitedefaultmidpunct}
{\mcitedefaultendpunct}{\mcitedefaultseppunct}\relax
\EndOfBibitem
\bibitem[Chithrananda \latin{et~al.}(2020)Chithrananda, Grand, and
  Ramsundar]{chemberta}
Chithrananda,~S.; Grand,~G.; Ramsundar,~B. ChemBERTa: Large-Scale
  Self-Supervised Pretraining for Molecular Property Prediction. 2020;
  \url{https://arxiv.org/abs/2010.09885}\relax
\mciteBstWouldAddEndPuncttrue
\mciteSetBstMidEndSepPunct{\mcitedefaultmidpunct}
{\mcitedefaultendpunct}{\mcitedefaultseppunct}\relax
\EndOfBibitem
\bibitem[Ahmad \latin{et~al.}(2022)Ahmad, Simon, Chithrananda, Grand, and
  Ramsundar]{chemberta2}
Ahmad,~W.; Simon,~E.; Chithrananda,~S.; Grand,~G.; Ramsundar,~B. ChemBERTa-2:
  Towards Chemical Foundation Models. 2022;
  \url{https://arxiv.org/abs/2209.01712}\relax
\mciteBstWouldAddEndPuncttrue
\mciteSetBstMidEndSepPunct{\mcitedefaultmidpunct}
{\mcitedefaultendpunct}{\mcitedefaultseppunct}\relax
\EndOfBibitem
\bibitem[Ho and Salimans(2022)Ho, and Salimans]{classifier-free}
Ho,~J.; Salimans,~T. Classifier-Free Diffusion Guidance. 2022;
  \url{https://arxiv.org/abs/2207.12598}\relax
\mciteBstWouldAddEndPuncttrue
\mciteSetBstMidEndSepPunct{\mcitedefaultmidpunct}
{\mcitedefaultendpunct}{\mcitedefaultseppunct}\relax
\EndOfBibitem
\bibitem[rdk()]{rdkit}
RDKit: Open-source cheminformatics. \url{https://www.rdkit.org}, Accessed:
  2024-06-19\relax
\mciteBstWouldAddEndPuncttrue
\mciteSetBstMidEndSepPunct{\mcitedefaultmidpunct}
{\mcitedefaultendpunct}{\mcitedefaultseppunct}\relax
\EndOfBibitem
\bibitem[Sun \latin{et~al.}(2020)Sun, Zhang, Banerjee, Bao, Barbry, Blunt,
  Bogdanov, Booth, Chen, Cui, \latin{et~al.} others]{pyscf}
Sun,~Q.; Zhang,~X.; Banerjee,~S.; Bao,~P.; Barbry,~M.; Blunt,~N.~S.;
  Bogdanov,~N.~A.; Booth,~G.~H.; Chen,~J.; Cui,~Z.-H.; others Recent
  developments in the PySCF program package. \emph{The Journal of chemical
  physics} \textbf{2020}, \emph{153}, 024109\relax
\mciteBstWouldAddEndPuncttrue
\mciteSetBstMidEndSepPunct{\mcitedefaultmidpunct}
{\mcitedefaultendpunct}{\mcitedefaultseppunct}\relax
\EndOfBibitem
\bibitem[Noutahi \latin{et~al.}(2024)Noutahi, Gabellini, Craig, Lim, and
  Tossou]{safe}
Noutahi,~E.; Gabellini,~C.; Craig,~M.; Lim,~J. S.~C.; Tossou,~P. Gotta be SAFE:
  a new framework for molecular design††Electronic supplementary
  information (ESI) available. See DOI: https://doi.org/10.1039/d4dd00019f.
  \emph{Digital Discovery} \textbf{2024}, \emph{3}, 796--804\relax
\mciteBstWouldAddEndPuncttrue
\mciteSetBstMidEndSepPunct{\mcitedefaultmidpunct}
{\mcitedefaultendpunct}{\mcitedefaultseppunct}\relax
\EndOfBibitem
\bibitem[Zhou \latin{et~al.}(2023)Zhou, Gao, Ding, Zheng, Xu, Wei, Zhang, and
  Ke]{uni-mol}
Zhou,~G.; Gao,~Z.; Ding,~Q.; Zheng,~H.; Xu,~H.; Wei,~Z.; Zhang,~L.; Ke,~G.
  Uni-Mol: A Universal 3D Molecular Representation Learning Framework. The
  Eleventh International Conference on Learning Representations. 2023\relax
\mciteBstWouldAddEndPuncttrue
\mciteSetBstMidEndSepPunct{\mcitedefaultmidpunct}
{\mcitedefaultendpunct}{\mcitedefaultseppunct}\relax
\EndOfBibitem
\bibitem[Zeng \latin{et~al.}(2023)Zeng, Li, and Li]{molkd}
Zeng,~L.; Li,~L.; Li,~J. MolKD: Distilling Cross-Modal Knowledge in Chemical
  Reactions for Molecular Property Prediction. 2023;
  \url{https://arxiv.org/abs/2305.01912}\relax
\mciteBstWouldAddEndPuncttrue
\mciteSetBstMidEndSepPunct{\mcitedefaultmidpunct}
{\mcitedefaultendpunct}{\mcitedefaultseppunct}\relax
\EndOfBibitem
\bibitem[Fang \latin{et~al.}(2022)Fang, Liu, Lei, He, Zhang, Zhou, Wang, Wu,
  and Wang]{gem}
Fang,~X.; Liu,~L.; Lei,~J.; He,~D.; Zhang,~S.; Zhou,~J.; Wang,~F.; Wu,~H.;
  Wang,~H. Geometry-enhanced molecular representation learning for property
  prediction. \emph{Nature Machine Intelligence} \textbf{2022}, \emph{4},
  127--134\relax
\mciteBstWouldAddEndPuncttrue
\mciteSetBstMidEndSepPunct{\mcitedefaultmidpunct}
{\mcitedefaultendpunct}{\mcitedefaultseppunct}\relax
\EndOfBibitem
\bibitem[Xia \latin{et~al.}(2022)Xia, Zhao, Hu, Gao, Tan, Liu, Li, and
  Li]{mole-bert}
Xia,~J.; Zhao,~C.; Hu,~B.; Gao,~Z.; Tan,~C.; Liu,~Y.; Li,~S.; Li,~S.~Z.
  Mole-bert: Rethinking pre-training graph neural networks for molecules. The
  Eleventh International Conference on Learning Representations. 2022\relax
\mciteBstWouldAddEndPuncttrue
\mciteSetBstMidEndSepPunct{\mcitedefaultmidpunct}
{\mcitedefaultendpunct}{\mcitedefaultseppunct}\relax
\EndOfBibitem
\bibitem[Qian \latin{et~al.}(2023)Qian, Tang, Yang, Liang, and Liu]{car}
Qian,~C.; Tang,~H.; Yang,~Z.; Liang,~H.; Liu,~Y. Can Large Language Models
  Empower Molecular Property Prediction? 2023;
  \url{https://arxiv.org/abs/2307.07443}\relax
\mciteBstWouldAddEndPuncttrue
\mciteSetBstMidEndSepPunct{\mcitedefaultmidpunct}
{\mcitedefaultendpunct}{\mcitedefaultseppunct}\relax
\EndOfBibitem
\bibitem[Honda \latin{et~al.}(2019)Honda, Shi, and Ueda]{smilestransformer}
Honda,~S.; Shi,~S.; Ueda,~H.~R. SMILES Transformer: Pre-trained Molecular
  Fingerprint for Low Data Drug Discovery. 2019;
  \url{https://arxiv.org/abs/1911.04738}\relax
\mciteBstWouldAddEndPuncttrue
\mciteSetBstMidEndSepPunct{\mcitedefaultmidpunct}
{\mcitedefaultendpunct}{\mcitedefaultseppunct}\relax
\EndOfBibitem
\bibitem[Chen \latin{et~al.}(2021)Chen, Gao, Nguyen, Chen, Jiang, Wei, and
  Pan]{agbt}
Chen,~D.; Gao,~K.; Nguyen,~D.~D.; Chen,~X.; Jiang,~Y.; Wei,~G.-W.; Pan,~F.
  Algebraic graph-assisted bidirectional transformers for molecular property
  prediction. \emph{Nature communications} \textbf{2021}, \emph{12}, 3521\relax
\mciteBstWouldAddEndPuncttrue
\mciteSetBstMidEndSepPunct{\mcitedefaultmidpunct}
{\mcitedefaultendpunct}{\mcitedefaultseppunct}\relax
\EndOfBibitem
\bibitem[Achiam \latin{et~al.}(2024)Achiam, Adler, Agarwal, Ahmad, Akkaya,
  Aleman, Almeida, Altenschmidt, Altman, Anadkat, \latin{et~al.}
  others]{chatgpt}
Achiam,~J.; Adler,~S.; Agarwal,~S.; Ahmad,~L.; Akkaya,~I.; Aleman,~F.~L.;
  Almeida,~D.; Altenschmidt,~J.; Altman,~S.; Anadkat,~S.; others GPT-4
  Technical Report. 2024; \url{https://arxiv.org/abs/2303.08774}\relax
\mciteBstWouldAddEndPuncttrue
\mciteSetBstMidEndSepPunct{\mcitedefaultmidpunct}
{\mcitedefaultendpunct}{\mcitedefaultseppunct}\relax
\EndOfBibitem
\bibitem[Sandfort \latin{et~al.}(2020)Sandfort, Strieth-Kalthoff, Kühnemund,
  Beecks, and Glorius]{mff}
Sandfort,~F.; Strieth-Kalthoff,~F.; Kühnemund,~M.; Beecks,~C.; Glorius,~F. A
  Structure-Based Platform for Predicting Chemical Reactivity. \emph{Chem}
  \textbf{2020}, \emph{6}, 1379--1390\relax
\mciteBstWouldAddEndPuncttrue
\mciteSetBstMidEndSepPunct{\mcitedefaultmidpunct}
{\mcitedefaultendpunct}{\mcitedefaultseppunct}\relax
\EndOfBibitem
\bibitem[Schwaller \latin{et~al.}(2020)Schwaller, Vaucher, Laino, and
  Reymond]{yield-bert-da}
Schwaller,~P.; Vaucher,~A.~C.; Laino,~T.; Reymond,~J.-L. Data augmentation
  strategies to improve reaction yield predictions and estimate uncertainty.
  2020;
  \url{https://chemrxiv.org/engage/chemrxiv/article-details/60c75258702a9b726c18c101}\relax
\mciteBstWouldAddEndPuncttrue
\mciteSetBstMidEndSepPunct{\mcitedefaultmidpunct}
{\mcitedefaultendpunct}{\mcitedefaultseppunct}\relax
\EndOfBibitem
\bibitem[Schwaller \latin{et~al.}(2021)Schwaller, Probst, Vaucher, Nair,
  Kreutter, Laino, and Reymond]{rxnfp}
Schwaller,~P.; Probst,~D.; Vaucher,~A.~C.; Nair,~V.~H.; Kreutter,~D.;
  Laino,~T.; Reymond,~J.-L. Mapping the space of chemical reactions using
  attention-based neural networks. \emph{Nature Machine Intelligence}
  \textbf{2021}, \emph{3}, 144--152\relax
\mciteBstWouldAddEndPuncttrue
\mciteSetBstMidEndSepPunct{\mcitedefaultmidpunct}
{\mcitedefaultendpunct}{\mcitedefaultseppunct}\relax
\EndOfBibitem
\bibitem[Loshchilov and Hutter(2019)Loshchilov, and Hutter]{adamw}
Loshchilov,~I.; Hutter,~F. Fixing Weight Decay Regularization in Adam. 2019;
  \url{https://arxiv.org/abs/1711.05101}\relax
\mciteBstWouldAddEndPuncttrue
\mciteSetBstMidEndSepPunct{\mcitedefaultmidpunct}
{\mcitedefaultendpunct}{\mcitedefaultseppunct}\relax
\EndOfBibitem
\bibitem[Paszke \latin{et~al.}(2019)Paszke, Gross, Massa, Lerer, Bradbury,
  Chanan, Killeen, Lin, Gimelshein, Antiga, Desmaison, Kopf, Yang, DeVito,
  Raison, Tejani, Chilamkurthy, Steiner, Fang, Bai, and Chintala]{pytorch}
Paszke,~A. \latin{et~al.}  PyTorch: An Imperative Style, High-Performance Deep
  Learning Library. Advances in Neural Information Processing Systems.
  2019\relax
\mciteBstWouldAddEndPuncttrue
\mciteSetBstMidEndSepPunct{\mcitedefaultmidpunct}
{\mcitedefaultendpunct}{\mcitedefaultseppunct}\relax
\EndOfBibitem
\end{mcitethebibliography}
\end{document}